\documentclass{ecai} 

\PassOptionsToPackage{table,dvipsnames}{xcolor}
\usepackage{xcolor}    
\usepackage{latexsym}
\usepackage{amssymb}
\usepackage{amsmath}
\usepackage{amsthm}
\usepackage{booktabs}
\usepackage{enumitem}
\usepackage{graphicx}
\usepackage{arydshln}
\usepackage{makecell}
\usepackage{colortbl}
\usepackage{hyperref}
\usepackage{cleveref}
\usepackage{subcaption}
\usepackage{tcolorbox}

\usepackage{xspace}
\newcommand{\ourmethod}{{\fontfamily{lmtt}\selectfont \textbf{AGP}}\xspace}
\newcommand{\llmname}[1]{{\fontfamily{pcr}\selectfont {#1}}\xspace}

\usepackage[normalem]{ulem}
\usepackage{bbm}
\usepackage{bbding}

\usepackage{pifont}         
\usepackage{units}
\newcommand{\fullsupp}{\(\checkmark\)}   
\newcommand{\nosupp}{\(\times\)}         
\newcommand{\partsupp}{\(\triangle\)}    

\definecolor{HeaderBlue}{HTML}{3478BF}
\definecolor{RowGrayA}{gray}{0.97}
\definecolor{RowGrayB}{gray}{0.90}
\definecolor{BrickRed}{HTML}{C0392B}
\definecolor{Goldenrod}{HTML}{DAA520}
\definecolor{SeaGreen}{HTML}{2E8B57}

\definecolor{ForestGreen}{RGB}{34,139,34}
\definecolor{myyellow}{RGB}{181, 181, 27}

\usepackage{enumitem}

\newcommand{\blue}[1]{$_{\color{BlueGreen}\downarrow #1}$}
\newcommand{\red}[1]{$_{\color{RedOrange}\uparrow #1}$}
\definecolor{darksalmon}{rgb}{0.91, 0.59, 0.48}
\definecolor{emerald}{rgb}{0.31, 0.78, 0.47}
\definecolor{green(pigment)}{rgb}{0.0, 0.65, 0.31}
\definecolor{amaranth}{rgb}{0.9, 0.17, 0.31}
\definecolor{iris}{rgb}{0.35, 0.31, 0.81}
\definecolor{uu}{rgb}{0.95, 0.51, 0.51}
\definecolor{spirodiscoball}{rgb}{0.06, 0.75, 0.99}

\usepackage{multirow}
\usepackage{makecell}
\usepackage{enumitem}
\usepackage{pifont}

\newtheorem{definition}{Definition}

\newcommand{\BibTeX}{B\kern-.05em{\sc i\kern-.025em b}\kern-.08em\TeX}
\usepackage{titlesec}
\usepackage{tocloft}
\titleformat{\paragraph}
  [runin]{\normalfont\normalsize\bfseries}{\theparagraph}{1em}{}[]

\titlespacing*{\paragraph}{0pt}{1ex plus .2ex minus .2ex}{1em}

\begin{document}


\begin{frontmatter}



\title{Adaptive Graph Pruning for Multi-Agent Communication}


\author[A]{\fnms{Boyi}~\snm{Li}}
\author[B]{\fnms{Zhonghan}~\snm{Zhao}\footnote{Project Leader. Email: zhaozhonghan@zju.edu.cn}}
\author[A]{\fnms{Der-Horng}~\snm{Lee}\footnotemark[*]}
\author[A, B]{\fnms{Gaoang}~\snm{Wang}\thanks{Corresponding Author. 
Email: \{dhlee, gaoangwang\}@intl.zju.edu.cn}}

\address[A]{Zhejiang University - University of Illinois Urbana-Champaign Institute}
\address[B]{Zhejiang University, College of Computer Science and Technology}


\begin{abstract} 
Large Language Model (LLM) based multi-agent systems have shown impressive performance across various fields of tasks, further enhanced through collaborative debate and communication using carefully designed communication topologies. However, existing methods typically employ a fixed number of agents or static communication structures, requiring manual pre-definition, and thus struggle to dynamically adapt the number of agents and topology simultaneously to varying task complexities. In this paper, we propose Adaptive Graph Pruning~(\ourmethod), a novel task-adaptive multi-agent collaboration framework that jointly optimizes agent quantity (hard-pruning) and communication topology (soft-pruning). Specifically, our method employs a two-stage training strategy: firstly, independently training soft-pruning networks for different agent quantities to determine optimal agent-quantity-specific complete graphs and positional masks across specific tasks; and then jointly optimizing hard-pruning and soft-pruning within a maximum complete graph to dynamically configure the number of agents and their communication topologies per task. Extensive experiments demonstrate that our approach is: \textbf{(1)~High-performing}, achieving state-of-the-art results across six benchmarks and consistently generalizes across multiple mainstream LLM architectures, with a increase in performance of $2.58\%\sim 9.84\%$; \textbf{(2)~Task-adaptive}, dynamically constructing optimized communication topologies tailored to specific tasks, with an extremely high performance in all three task categories (general reasoning, mathematical reasoning, and code generation); \textbf{(3)~Token-economical}, having fewer training steps and token consumption at the same time, with a decrease in token consumption of $90\%+$; and \textbf{(4)~Training-efficient}, achieving high performance with very few training steps compared with other methods. The performance will surpass the existing baselines after about ten steps of training under six benchmarks. Our code and demos are publicly available at~\url{https://resurgamm.github.io/AGP/}.
\end{abstract}

\end{frontmatter}
s
\vspace{-3mm}
\section{Introduction}
LLM-based agents have advanced artificial intelligence by integrating language generation with decision-making and action execution. These agents excel across diverse applications, including code generation~\citep{meta-gpt}, embodied agent~\citep{voyager, zhao2023see, zhao2025rig}, \textit{etc.}, with a great performance improvement over before. However, the inability of single-agent systems to leverage collective intelligence and foster collaboration limits their effectiveness in solving the intricate tasks present in real-world scenarios. Notably, the collaborative potential of multiple LLM agents, structured within carefully designed communication topologies, has demonstrated superior performance compared to single-agent systems, highlighting the critical role of communication structure in multi-agent intelligence~\citep{reflexion, autogen, hu2024magraph}. This has led the current research focus to multi-agent collaborative systems, in which multiple agents engage in iterative interactions within a shared environment, regarding other agents as additional elements of their environment, and continuously refining their strategies and answers through learning from different perspectives of other agents. This process allows them to collectively pursue and achieve a common goal~\citep{hu2024magraph}.

\begin{figure}[!t]
  \centering
  \includegraphics[width=1\linewidth]{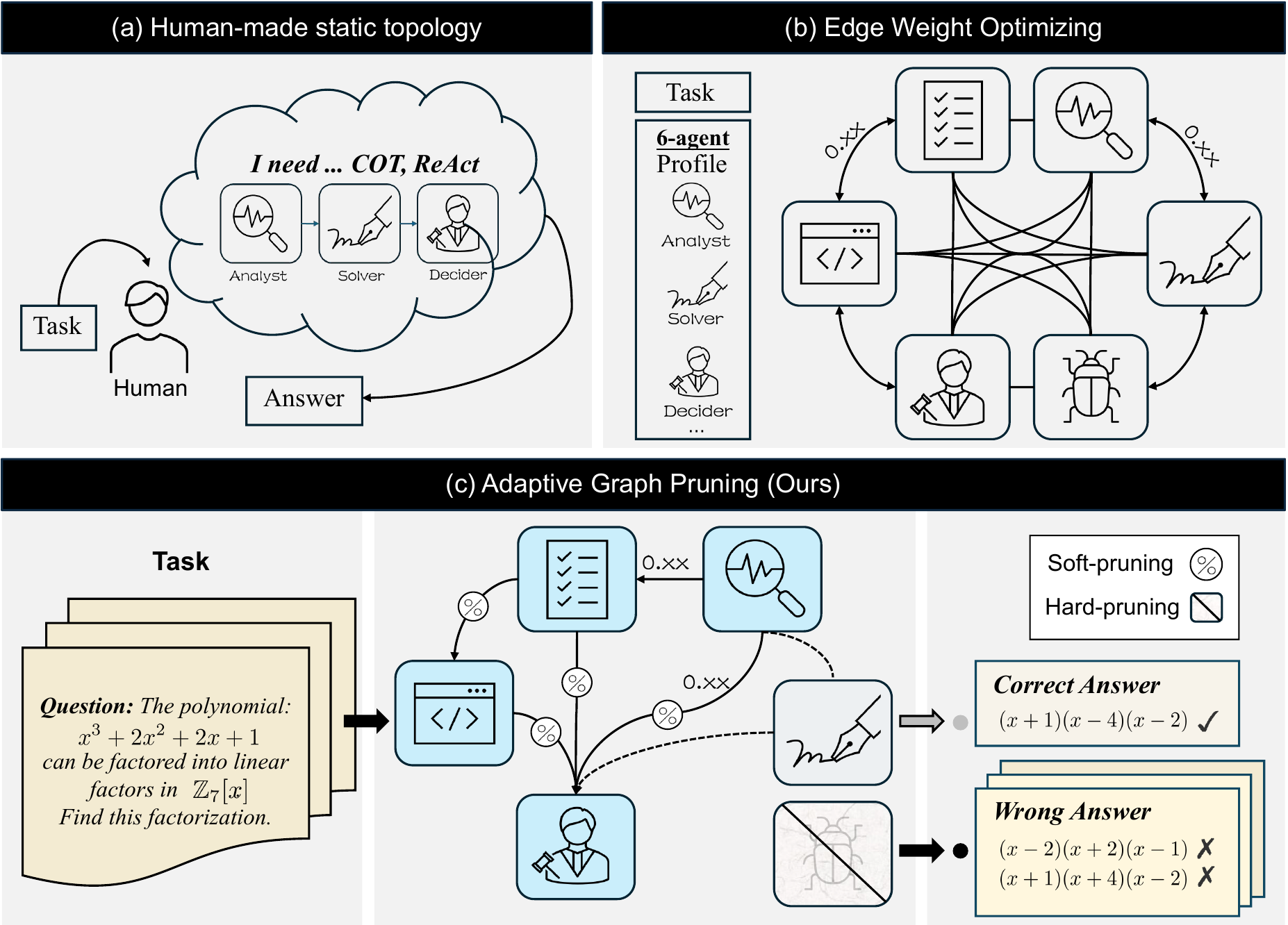}
  \caption{Comparison between existing works and \ourmethod. \ourmethod produces a dual-pruning method to generate task-adaptive communication topologies.}
   \label{fig:teaser}
   \vspace{4mm}
\end{figure}

Despite rapid progress, selecting a communication topology that truly suits each task remains challenging. Most studies still rely on \textit{human‑made, static} layouts like chains~\citep{cot,meta-gpt}, trees~\citep{tot,autogen, zhao2024hierarchical, zhao2024we}, stars~\citep{autogen}, complete graphs, or random graphs~\citep{qian2024scaling} (\Cref{fig:teaser},(a)). These designs embed strong human assumptions: they may succeed in one domain yet fail in another~\citep{zhuge2024gptswarm, zhang2024g}. Moreover, per‑task manual tuning is costly and often overlooks not intuitively relevant agents.
More recent work relaxes this constraint by \textit{optimizing edge weights} within a fixed agent pool~\citep{zhuge2024gptswarm, zhang2024g} (\Cref{fig:teaser},(b)). Despite the improvement, the topology remains constrained by a fixed number of predefined agents, which may overlook intuitively relevant agents and limit flexibility and scalability.

We instead pursue a \emph{fully task‑adaptive} topology. As \Cref{fig:teaser},(c) shows, activating a key agent (Math Solver) while pruning an irrelevant one (Insect Researcher) yields higher quality in the math domain at lower cost. Experiments further reveal that the most useful agents are sometimes counter‑intuitive, underscoring the limits of human design (\Cref{sec:exp_task_ad}). To realize this vision, we propose \textbf{Adaptive Graph Pruning} (termed as \ourmethod), a task‑adaptive collaboration framework that jointly optimizes agent quantity (hard pruning) and their communication topology (soft pruning). Because task complexity shapes both the optimal number of agents and their interaction pattern, an adaptive and dynamic topology is essential for consistent, scalable performance.

Our proposed approach incorporates two stages. In \textbf{Stage I}, we sample various communication topologies from the agent pool we collected and utilize a graph auto-encoder (GAE) as the soft-pruning module to get the optimal communication topology for the corresponding type of task, composing all the tasks and the corresponding communication topologies as data pairs to obtain the dataset. In \textbf{Stage II}, we add the hard-pruning module that shares the same latent space with the soft-pruning module, jointly optimizing hard-pruning and soft-pruning within a maximum complete graph by calculating the relative loss with the task-communication-topology data pair, and finally dynamically configuring the number of agents and their communication topologies per task.

We conduct extensive experiments across six benchmark tasks with twelve baselines, demonstrating that our approach achieves state-of-the-art performance with an average performance improvement of $2.58\%\sim 9.84\%$. Moreover, the comparison with the baselines on training steps, token quantity consumption, and accuracy rate can illustrate that our method is simultaneously training-efficient and token-economical. With only a ten-step training, \ourmethod achieves more than a $90\%$ decrease of token consumption in prompting as well as an outstanding performance.
Our contribution can be summarized as follows:

\begin{itemize}

\item A novel task-adaptive multi-agent collaboration framework, dynamically constructing optimized communication topologies tailored specifically to individual tasks.

\item A corresponding two-stage training strategy, jointly optimizing agent quantity (hard-pruning) and communication topology (soft-pruning) for \ourmethod.

\item Our method delivers state-of-the‑art performance while offering excellent inference token economy and high training efficiency.

\end{itemize}

\section{Related Works}\label{sec:related}

\paragraph{LLM Agent Collaboration Topologies.} Early work established the superiority of single-agent LLMs for reasoning and planning. These works prompt single-agent LLMs in a designed way, such as chain of thought (CoT)~\citep{cot}, tree of thought (ToT)~\citep{tot}, graph of thought (GoT)~\citep{got}, complexity-based prompting~\citep{fu2022complexity}, and self-consistency (SC)~\citep{wang2023selfconsistency}, to improve text-based reasoning. Subsequent studies showed that \emph{multiple} LLM agents can outperform a single model by combining specialized skills, from majority voting~\cite{chen2024compundLLM} to more intricate interaction schemes~\cite{chen2023agentverse}. To have further enhanced performance and better agent integration capabilities, recent works explore diverse \emph{pre-defined} or \emph{static} communication topologies:  
\textbf{(1)~non-interactive} designs in which agents operate alone without any interactions with others, such as LATM~\citep{zhang2023astools}, LLM-Blender\citep{blender}, and LLM-Debate~\citep{arXiv2023_MultiAgent-Debate};  
\textbf{(2)~chain} structures~\citep{qian2024scaling} in ChatDev~\citep{software-dev}, MetaGPT~\citep{meta-gpt}, and L2MAC~\citep{holt2024l2mac}, where each agent receiving and passing information one-by-one;  
\textbf{(3)~star} patterns~\citep{qian2024scaling} with a central commander or manager that manages other agents in AutoGen~\citep{autogen} and MiniGrid~\citep{zhou2023large};  
\textbf{(4)~tree} hierarchies~\citep{qian2024scaling} with a root agent that manages all child agents, such as SoA~\cite{ishibashi2024selforganize-mother}; and  
\textbf{(5)~general graphs} including complete or random variants~\cite{qian2024scaling}. \textbf{(6)~fixed nodes} such as GPTSwarm\citep{zhuge2024gptswarm} and G-Designer~\cite{zhang2024g}, which optimize edge weights but are not sensitive to the number of nodes, limiting their task adaptation capabilities.
However, our Adaptive Graph Pruning~(\ourmethod) is an \emph{adaptive} topology learner that \emph{jointly} soft-prunes edges and hard-prunes nodes, yielding a task-specific graph that can even shrink below the agent pool size to lower communication cost while boosting accuracy.

\paragraph{Automating Agentic Systems.} Research on automating the design of agent‑based systems can be grouped into three threads: \textbf{(1) Prompt Optimization}, exemplified by PromptBreeder \citep{fernando2023promptbreeder}, DsPy \citep{khattab2023dspy}, and EvoPrompt \citep{guo2023evoprompt}; \textbf{(2) Agent Profiling}, including AgentVerse \citep{chen2023agentverse}, EvoAgent \citep{yuan2024evoagent}, and AutoAgents \citep{chen2023autoagents}; and \textbf{(3) Inter‑agent Communication}, which orchestrates the information flow among agents, as explored by GPTSwarm \citep{zhuge2024gptswarm}, DyLAN \citep{arXiv2023_Dynamic-LLM-Agent}, EvoMAC \citep{hu2024evomac}, AgentPrune \citep{zhang2024cut}, and G‑Designer \citep{zhang2024g}.  A recent surge of work pushes this line further by employing search or evolutionary algorithms to explore ever‑larger system spaces. For instance, ADAS \citep{hu2024adas} and AgentSquare \citep{shang2024agentsquare} automate single‑agent design, while AFlow \citep{zhang2024aflow} uses Monte Carlo tree search for multi‑agent workflow generation, and MaAS \citep{zhang2025multi} searches architectural distributions. Although such approaches can yield strong performance, their freedom to explore vast configuration spaces incurs substantial computational overhead. By contrast, \ourmethod\ sidesteps costly global searches: it adapts communication topologies on the fly over the complete graph, composing task‑specific subgraphs that deliver higher efficiency and lower average training and inference costs (\textit{e.g.}, token usage) without sacrificing performance.

\paragraph{Communication Topologies as Graphs.} Graphs are a natural abstraction for agent communication and have long been used in MARL~\cite{pesce2023learning,hu2024magraph}.  
With the advent of LLM agents, researchers recognize that the way multiple agents communicate can be considered from the perspective of the graph, regarding communication topologies as graphs. This approach helps to illustrate the connections between the agents effectively. ~\citep{hu2024magraph}. This idea is then used for implicit graphs like ChatEval~\citep{chateval}, AutoGen~\citep{autogen}, and DSPy~\cite{khattab2023dspy}.  
Recent works move further to \emph{explicit} graph representations: ChatLLM~\citep{chatllm-network} and DyLAN~\citep{arXiv2023_Dynamic-LLM-Agent} adopt layered MLP-like graphs; MacNet~\citep{qian2024scaling} designs various human-made structures~\cite{qian2024scaling}; GPTSwarm~\citep{zhuge2024gptswarm} and G-Designer~\citep{zhang2024g} regard communication topologies as learnable graphs, designing communication topologies with a fixed number of nodes by just optimizing edge weights. 
Existing graphs are either predefined or globally optimized once, thus ignoring the per-task difficulty and the heterogeneous usefulness of agents. Our Adaptive Graph pruning~(\ourmethod), leveraging a \emph{graph pool} collected by the collector, we derive per-task ground-truth sub-graphs and train a dual-pruning GNN that \textit{dynamically} tailors both connectivity and agent subset: the first approach to deliver fully \emph{task-adaptive}, \emph{size-variable} LLM-agent topologies.

\section{Method}
\label{sec:method}

\begin{figure*}[!t]
  \centering
  \includegraphics[width=\linewidth]{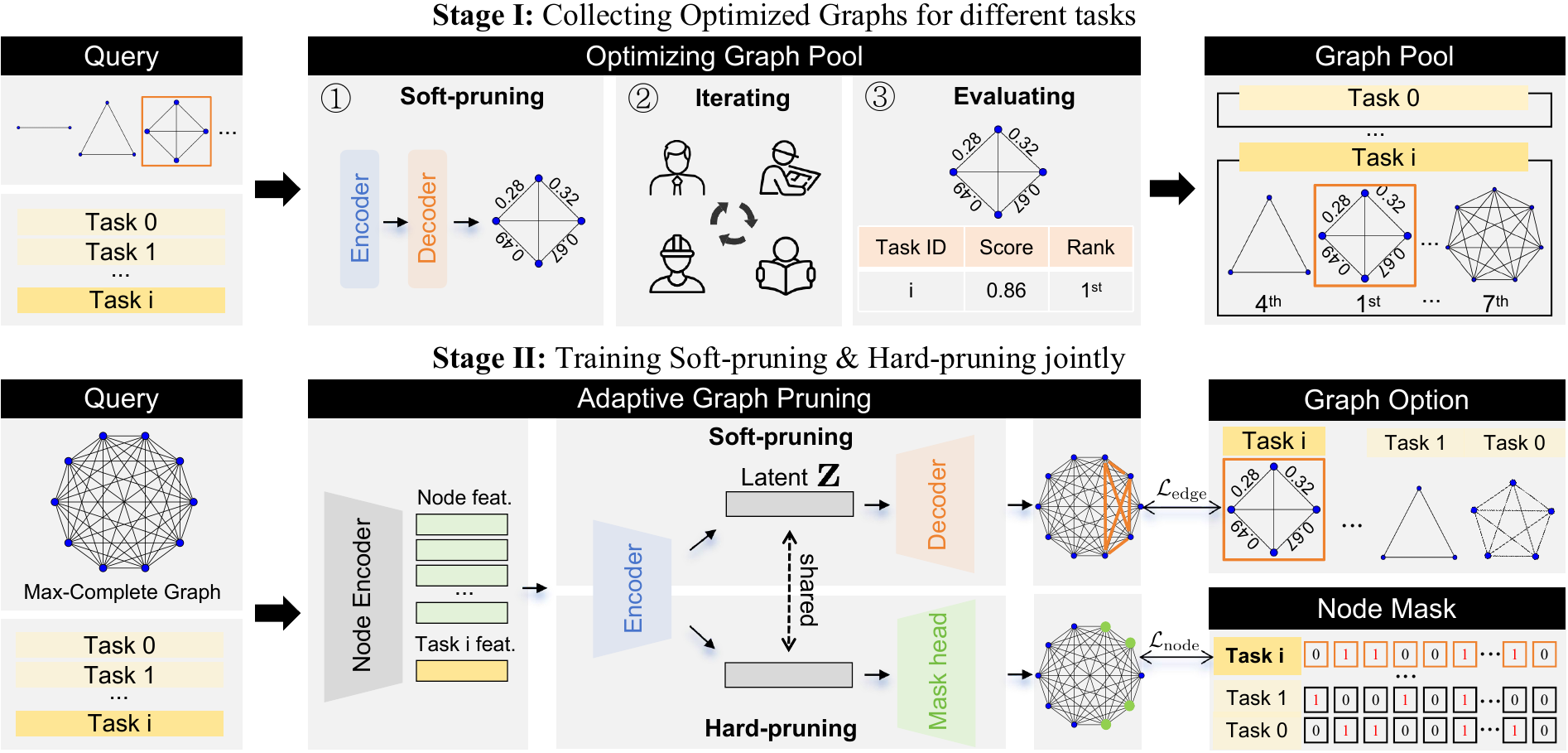}
  \caption{\textbf{\ourmethod} (\textit{Adaptive Graph Pruning}) framework.  
  Stage I mines near‑optimal sub‑graphs from a heterogeneous agent pool;  
  Stage II trains a joint soft–/hard–pruning network that instantiates an
  adaptive communication topology for any incoming query~$\mathcal{Q}$.}
  \label{fig:framework}
  \vspace{8pt}
\end{figure*}

Adaptive Graph Pruning (termed as \ourmethod) first mines high‑utility sub‑graphs from a fixed pool of heterogeneous LLM‑agents and preserves their edge labels and node masks as supervision. The next section sets up notation, casts these ideas in graph‑topological terms, and details the multi‑agent communication protocol that \ourmethod learns to instantiate for any new query. All notation is listed in Appendix A.2 of \cite{li2025adaptive}.

\subsection{Preliminaries}
\label{sec:prelim}

\paragraph{Multi‑agent system.} We represent a set of $N_{\max}$ large‑language‑model (LLM) agents as a directed graph $\mathcal{G}=(\mathcal{V},\mathbf{A})$ with $\mathcal{V}=\{v_1,\dots,v_{N_{\max}}\}$ and $\mathbf{A}\!\in\!\{0,1\}^{N_{\max}\times N_{\max}}$, where $A_{ij}=1$ if and only if messages may flow from $v_i$ to~$v_j$. Each agent is a tuple:
\begin{equation}
  v_i=\bigl\langle
      \mathrm{LM}_i,\;
      r_i,\;
      s_i,\;
      \phi_i
  \bigr\rangle,
\end{equation}
which contains its language backbone $\mathrm{LM}_i$, role/persona $r_i$,  dialogue state (history) $s_i$, and optional tool plug‑ins $\phi_i$ (\textit{e.g.}, Python, Wolfram).

\paragraph{Design objective.}
Given a query $\mathcal{Q}$, the team communicates for
$K$ rounds and outputs $a^{(K)}$.
We aim to select a communication topology $\mathbf{A}$ that maximizes utility while minimizing inter‑agent traffic from the sub-graph set $\mathfrak{G}$ of the maximum complete graph with $N_{\max}$ nodes:
\begin{equation}
\label{eq:objective}
\min_{\mathbf{A}\in\mathfrak{G}}
\Bigl[
  -U(\mathbf{A}\mid\mathcal{Q})
  +\lambda_c\,C(\mathbf{A})
\Bigr],
\end{equation}
where $U$ is task utility
(\textit{e.g.}, exact‑match accuracy),
$C$ counts the number of transmitted tokens, and
$\lambda_c\!>\!0$ balances the two.

\paragraph{Two degrees of freedom.}
\begin{definition}[\textbf{Soft‑pruning}]
Assign each potential edge a weight
$w_{ij}\!\in\![0,1]$,
yielding a weighted adjacency
$\mathbf{W}$; small $w_{ij}$ throttles or blocks the
bandwidth of $(v_i\!\to v_j)$.
\end{definition}
\begin{definition}[\textbf{Hard‑pruning}]
Introduce a binary mask
$\mathbf{m}\!\in\!\{0,1\}^{N_{\max}}$ where
$m_i=0$ removes agent $v_i$ from the task, producing a node‑induced
sub‑graph $\mathcal{G}[\mathbf{m}]$.
\end{definition}

\noindent \ourmethod jointly learns $\mathbf{W}$ and $\mathbf{m}$
so that the resulting graph
$\mathcal{G}_{\mathrm{com}}=\mathcal{G}[\mathbf{m}],\mathbf{W}$
is both task‑aware and cost‑efficient.

\subsection{Stage I: Collecting Optimized Graphs}
\label{sec:collector}

\Cref{fig:framework} (higher half) shows how \ourmethod samples various communication topologies from the agent pool, collects optimized graphs from the sub-graph set of the maximum complete graph, and finally gets the graph pool. Specifically, \textbf{Stage I} contains the following parts:

\paragraph{Agent indexing.}
We permanently anchor the $N_{\max}$ heterogeneous agents to the
complete graph $K_{N_{\max}}$, eliminating permutation ambiguity.

\paragraph{Sampling pool.}
For every order $i\!\in\![2,N_{\max}]$
the \textsc{Collector} samples a complete sub‑graph $K_i$ by drawing
a size‑$i$ agent subset uniformly without replacement
and inserting them in ascending ID order.
Graph orders follow a truncated Gaussian
$\mathcal{N}(\mu=N_{\max}/2,\sigma^{2})$
until a pool budget $B$ is met.\footnote{%
A typical configuration uses $B=2{,}000$, $\sigma=2$.}

\paragraph{Task training \& scoring.}
Each sampled graph $\mathcal{G}$ is fine‑tuned on its associated task
$t$ to obtain
$u_t(\mathcal{G})=\mathrm{Acc}(\mathcal{G};t)\!\in\![0,1]$.
Per task, we keep the top‑2 performers
$\bigl\{\mathcal{G}^{(1)}_{t},\mathcal{G}^{(2)}_{t}\bigr\}$
and \emph{lift} them to the
$N_{\max}$‑node reference frame:
\begin{equation}
  \textstyle
  \mathbf{A}^{\mathrm{gt}}\!\in\!\{0,1\}^{N_{\max}\times N_{\max}},\quad
  \mathbf{y}\!\in\!\{0,1\}^{N_{\max}}.
\end{equation}
The pair $(\mathbf{A}^{\mathrm{gt}},\mathbf{y})$ acts as exact supervision for Stage II.

\paragraph{Corpus statistics.}
The resulting 460 supervision graphs span three
families (\Cref{fig:composition}):
200 \emph{General Reasoning},
100 \emph{Mathematical Reasoning},
and 160 \emph{Code Generation}.

\subsection{Stage II: Training Soft‑pruning \& Hard‑pruning}
\label{sec:topology}
\Cref{fig:framework} (lower half) shows how \textbf{Stage II}
turns the \emph{static} supervision pool harvested in Stage I
into a \emph{dynamic} topology generator that adapts to every query.
Conceptually, the module follows a two–branch design:

\begin{enumerate}
\item \textbf{Soft‑pruning path} (orange in the figure) learns a
      \emph{directed, weighted} adjacency that modulates
      \underline{how much} information flows between any two retained
      agents.
\item \textbf{Hard‑pruning path} (green) learns a binary node mask that
      decides \underline{which} agents are retained at all.
\end{enumerate}

\noindent Both paths share a common latent representation~$\mathbf{Z}$, so the edge weights and node decisions remain mutually consistent.

\vspace{0.4em}
\subsubsection{Network Architecture}

To satisfy our conception, the network architecture of \ourmethod consists of four modules: a node encoder to embed the agent profile and task into the network; a GCN backbone to calculate the features on the maximum complete graph; an edge-weight head to generate an edge weight matrix for the maximum complete graph; and a node-mask head to generate mask matrix for nodes. Here are the detailed description of these modules.

\paragraph{Node encoder.}
Each agent profile and the task‑specific virtual node
(see Stage I) are embedded into $\mathbf{X}\!\in\!\mathbb{R}^{N_{\max}\times d}$
via a lightweight Sentence‑BERT encoder, giving the model access to both role/tool metadata and query semantics.

\paragraph{GCN backbone.}
A two‑layer Graph Convolutional Network
aggregates neighborhood information over the \emph{max‑complete} graph,
producing latent vectors
$\mathbf{Z}=\mathrm{GCN}(\mathbf{X})\in\mathbb{R}^{N_{\max}\times h}$.
This step corresponds to the \emph{Node feat.} and
\emph{Latent $\mathbf{Z}$} blocks in \Cref{fig:framework}.

\paragraph{Edge‑weight head.}
A bilinear projector
$\mathrm{Proj}_{\mathrm{edge}}\!: \mathbb{R}^{h}\times\mathbb{R}^{h}\!\to\![0,1]$
maps every ordered pair $(\mathbf{z}_i,\mathbf{z}_j)$ to
$W^{\mathrm{pred}}_{ij}$, yielding a dense,
directed weight matrix
$\mathbf{W}^{\mathrm{pred}}\in[0,1]^{N_{\max}\times N_{\max}}$
with zero diagonal.

\paragraph{Node‑mask head.}
A two‑layer MLP produces logits
$\mathbf{s}\in\mathbb{R}^{N_{\max}}$ which, after a sigmoid,
become continuous masks
$\hat{\mathbf{y}}=\sigma(\mathbf{s})$.
During inference we threshold $\hat{y}_i\!\ge\!0.5$ to keep
only the most relevant agents, as illustrated by the ``Node Mask''
panel in \Cref{fig:framework}.

\vspace{0.4em}
\subsubsection{Training Details}

\paragraph{Why two losses?}
Edge supervision alone cannot shrink a graph
(all nodes remain active), while node‑only sparsification collapses the topology into a clique (without edge weighting).
Our design, therefore, couples \emph{edge loss}
to learn bandwidths and \emph{node loss}
to learn minimal agent subsets.

\paragraph{Edge loss (soft‑pruning).}

As shown in below \Cref{eq:edge-loss}, we measure MSE on positive edges (first term)
and pushes negatives towards 0 (second term),
discounting edges whose endpoints are masked out by~$\mathbf{y}$.

\begin{align}
\mathcal{L}_{\text{edge}}
=&\;
\frac{1}{\lVert\mathbf{M}\rVert_{1}}
\sum_{\substack{i\neq j}} M_{ij}\,
\mathrm{MSE}\!\bigl(W^{\text{pred}}_{ij},A^{\text{gt}}_{ij}\bigr)
\notag\\
&+
\lambda_{\text{off}}\,
\frac{1}{n(n-1)-\lVert\mathbf{M}\rVert_{1}}
\sum_{\substack{i\neq j}} (1\!-\!M_{ij})\,{W^{\text{pred}}_{ij}}^{2},
\label{eq:edge-loss}
\end{align}
where $M_{ij}=y_{i}y_{j}$ masks out edges whose
\emph{either} endpoint is absent in the ground‑truth subgraph.

\paragraph{Node loss (hard‑pruning).}

As shown in below \Cref{eq:node-loss}, we measure BCE between the ground-truth mask matrix and the mask matrix generated. Specifically, let $\hat{\mathbf{y}}=\sigma(\mathbf{s})$:
\begin{align}
  \mathcal{L}_{\text{node}}
  =\mathrm{BCE}(\hat{\mathbf{y}},\mathbf{y})
   +\lambda_{s}\bigl\langle\hat{\mathbf{y}}\bigr\rangle
   +\lambda_{c}\frac{1}{n^{2}}
     \sum_{i\notin\mathbf{y}}\sum_{j}|W^{\text{pred}}_{ij}|. 
\label{eq:node-loss}
\end{align}
where the BCE term aligns predicted masks with ground-truth and the sparsity penalty $\lambda_s\langle\hat{\mathbf{y}}\rangle = \frac{1}{n}\sum_{i=1}^{n}\hat{y}_i$. The sparsity penalty $\lambda_s\langle\hat{\mathbf{y}}\rangle$ encourages fewer active agents, while the coherence term penalizes any outgoing weight from ground‑truth \emph{absent} nodes, ensuring the two branches agree.

\paragraph{Total objective.}\label{sec:tot-obj}
$\mathcal{L}_{\text{total}}
=\mathcal{L}_{\text{edge}}
+\beta\,\mathcal{L}_{\text{node}}$
balances structural fidelity and sparsity.

\vspace{0.4em}

\paragraph{Continuous–Discrete bridge.}
Both $W^{\mathrm{pred}}_{ij}$ and $\hat{y}_i$ use the Gumbel–Sigmoid trick so gradients flow through the discrete sampling process. We anneal at the temperature $\tau{:}1.0\!\rightarrow\!0.1$, so early epochs explore many graphs, while late epochs commit to crisp 0/1 decisions, mirroring the \textbf{blurry $\to$ sharp} transition sketched in the ``Soft‑pruning'' panel of \Cref{fig:framework}.

\subsection{Dataset Composition}
\label{sec:data}

\begin{table*}[!t]
  \centering
  \caption{Performance comparison with single‑agent approaches, multi‑agent topologies, and \ourmethod.
           The base LLM for all baselines is \llmname{gpt‑4o‑mini}. Please refer to Appendix A.3~\cite{li2025adaptive} for \llmname{gpt-3.5-turbo} performance.
           We \textbf{bold} the best results and \underline{underline} the runner‑ups.
           “Mul.” and “Ada.” indicate multi‑agent support and task adaptivity, respectively.
           \nosupp, \partsupp, and \fullsupp denote no, partial, and full support.}
  \label{tab:rq1_performance}

  \setlength{\tabcolsep}{5.3pt}
  \renewcommand\arraystretch{1.15}

  \rowcolors{2}{gray!10}{white}

  \footnotesize
  \resizebox{\linewidth}{!}{%
  \begin{tabular}{lccllllllll}   
    \toprule
    \rowcolor{gray!25}
    \textbf{Method} & \textbf{Mul.} & \textbf{Ada.} & \textbf{MMLU} & \textbf{GSM8K} &
    \textbf{MultiArith} & \textbf{SVAMP} & \textbf{AQuA} & \textbf{HumanEval} & \textbf{Avg.} \\
    \midrule
    Single‑agent                          & \nosupp  & \nosupp  & 77.81 & 87.45 & 96.85 & 88.26 & 71.42 & 87.08 & 84.81\\
    CoT~\citep{cot}                       & \nosupp  & \nosupp  & 78.43\red{0.62} & 87.10\blue{0.35} & 96.31\blue{0.54} & 86.24\blue{2.02} & 65.00\blue{6.42} & 88.13\red{1.05} & 83.54\\
    ComplexCoT~\citep{fu2022complexity}   & \nosupp  & \nosupp  & 81.05\red{3.24} & 86.89\blue{0.56} & 96.70\blue{0.15} & 90.53\red{2.27} & 77.94\red{6.52} & 87.49\red{0.41} & 86.77\\
    SC (CoT$\times$5)~\citep{wang2023selfconsistency}
      & \nosupp  & \nosupp  & 80.96\red{3.15} & 87.57\red{0.12} & 96.58\blue{0.27} & 87.92\blue{0.34} & 70.90\blue{0.52} & 88.60\red{1.52} & 85.42\\
    \midrule
    MacNet~\citep{qian2024scaling}        & \fullsupp & \nosupp & 82.98\red{5.17} & 87.95\red{0.50} & 96.03\blue{0.82} & 88.06\blue{0.20} & 73.20\red{1.78} & 84.57\blue{2.51} & 85.47\\
    AgentVerse~\citep{chen2023agentverse} & \fullsupp & \nosupp & 78.36\red{0.55} & 89.91\red{2.46} & 97.50\red{0.65} & 89.64\red{1.38} & 76.41\red{4.99} & 89.29\red{2.21} & 86.85\\
    MetaGPT~\citep{meta-gpt}              & \fullsupp & \nosupp & -- & -- & -- & -- & -- & \underline{90.93}\red{3.85} & --\\
    LLM‑Blender~\citep{blender}           & \fullsupp & \nosupp & 81.22\red{3.41} & 88.35\red{0.90} & 97.29\red{0.44} & 89.52\red{1.26} & 77.28\red{5.86} & 88.80\red{1.72} & 87.08\\
    LLM‑Debate~\citep{arXiv2023_MultiAgent-Debate}
      & \fullsupp & \nosupp & 81.04\red{3.23} & 89.47\red{2.02} & 97.33\red{0.48} & \underline{91.76}\red{3.50} & 78.60\red{7.18} & 88.68\red{1.60} & 87.81\\
    DyLAN~\citep{arXiv2023_Dynamic-LLM-Agent}
      & \fullsupp & \partsupp & 79.96\red{2.15} & 89.98\red{2.53} & 97.12\red{0.27} & 88.48\red{0.22} & 75.11\red{3.69} & 90.42\red{3.34} & 86.85\\
    GPTSwarm~\citep{zhuge2024gptswarm}    & \fullsupp & \partsupp & 82.80\red{4.99} & 89.14\red{1.69} & 96.79\blue{0.06} & 87.02\blue{1.24} & 78.40\red{6.98} & 89.32\red{2.24} & 87.25\\
    G‑Designer~\cite{zhang2024g}          & \fullsupp & \partsupp & \underline{87.20}\red{9.39} & \underline{93.97}\red{6.52} & {98.33}\red{1.48} & 90.29\red{2.03} & \underline{80.07}\red{8.65} & 87.50\red{0.42} & 89.56\\
    MaAS~\cite{zhang2025multi} & \fullsupp & \partsupp & -- & 92.30\red{4.85} & \underline{98.80}\red{1.95} & -- & -- & \textbf{92.85}\red{5.77} & -- \\
    \midrule
    \ourmethod{} (Ours)                  & \fullsupp & \fullsupp & \textbf{87.65}\red{9.84} & \textbf{95.01}\red{7.56} & \textbf{99.44}\red{2.58} & \textbf{92.30}\red{4.04} & \textbf{81.20}\red{9.78} & {90.62}\red{3.54} & \textbf{91.04}\\
    \bottomrule
  \end{tabular}}
\end{table*}

\begin{figure}[t]
  \centering
  \includegraphics[width=\linewidth]{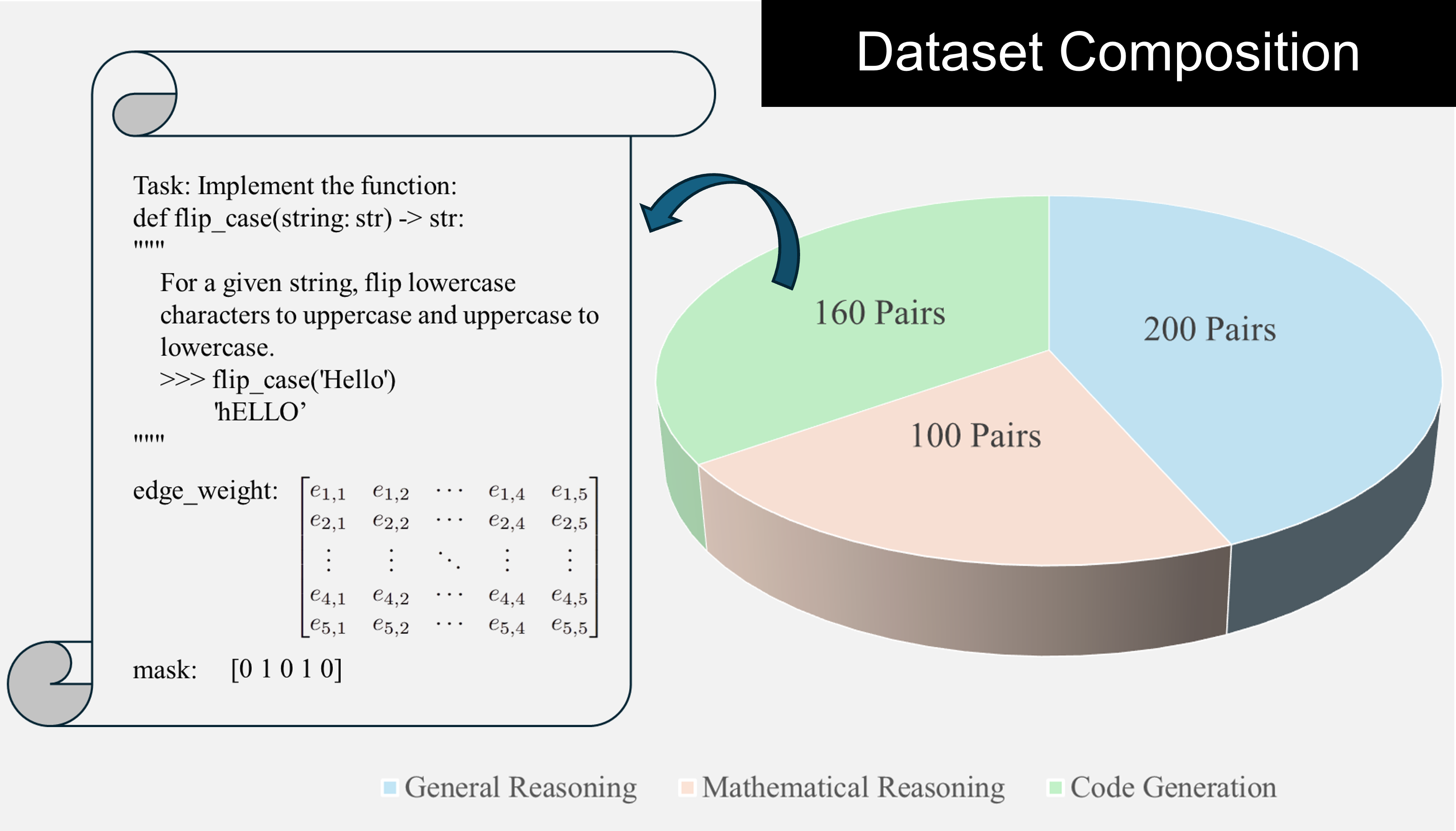}
  \vspace{0.01mm}
  \caption{\textbf{Stage I} supervision corpus.  
           \textit{Left:} one labeled pair comprising (i) a natural‑language
           task description, (ii) the edge‑weight matrix of the
           \emph{max‑complete} graph ($e_{i,j}$ is the weight of
           $v_i\!\rightarrow\!v_j$), and (iii) the binary node‑mask
           vector.  \textit{Right:} category breakdown of all
           460 pairs.}
  \label{fig:composition}
  \vspace{5mm}
\end{figure}

\Cref{fig:composition} visualizes both the \emph{micro} and
\emph{macro} structure of our supervision corpus. More data statistics are listed in Appendix A7~\cite{li2025adaptive}.

\paragraph{Basic unit (left panel).}
Each labeled pair is a triple
\(\langle \text{Task},\,\mathbf{A}^{\mathrm{gt}},\,\mathbf{y}\rangle\):

\begin{itemize}
  \item \textbf{Task description}\,:\,natural‑language prompt that an
        LLM team must solve (\textit{e.g.}\ implement `flip\_case`).
  \item \textbf{Edge‑weight matrix}
        \(\mathbf{A}^{\mathrm{gt}}\in\mathbb{R}^{N_{\max}\times N_{\max}}\)\,:\,
        the fully specified communication graph lifted to $K_{N_{\max}}$;
        entry \(e_{i,j}\) encodes the ground‑truth importance of
        directed edge $(v_i\!\to v_j)$.
  \item \textbf{Node mask}
        \(\mathbf{y}\in\{0,1\}^{N_{\max}}\)\,:\,identifies agents that
        actually participate in the task.
\end{itemize}

\noindent These rich labels drive the joint soft‑/hard‑pruning learner described
in \Cref{sec:topology}.

\paragraph{Global statistics (right panel).}
The corpus totals \textbf{460} supervision graphs, drawn from three
domains:

\begin{itemize}

  \item \textbf{General Reasoning}  \textit{(200 pairs)}: commonsense QA,
        humanities, social sciences; serves to anchor broad
        language understanding.
  \item \textbf{Mathematical Reasoning} \textit{(100 pairs)}: arithmetic,
        algebra, and word‑problem reasoning; stresses symbolic
        manipulation and multi‑step deduction.
  \item \textbf{Code Generation} \textit{(160 pairs)}: implement functions
        from I/O specifications or docstrings; examines the ability
        to plan, execute, and self‑verify algorithmic tasks.
\end{itemize}

\paragraph{Why this mix?}
Combining natural‑language, numerical, and programmatic tasks forces the topology learner to \emph{generalize} beyond single‑domain heuristics: agents required for math (\textit{e.g.}\ a “Math Solver”) differ from those for code generation(\textit{e.g.}\ a “Programming Expert”), while general reasoning often benefits from broad, low‑bandwidth exchanges.  The diverse 460‑pair corpus, therefore, supplies both breadth and depth to train a robust, query‑adaptive pruning policy.

\section{Experiments}\label{sec:exp}

We evaluate \textbf{\ourmethod} on six public benchmarks that cover three task families: general knowledge, mathematical reasoning, and code generation. The comparison set includes the strongest single‑agent prompting techniques (CoT, Complex‑CoT, Self‑Consistency) and eight multi‑agent systems ranging from fixed graphs to adaptive designs and architecture‑search. 

\begin{figure*}[!h]
  \centering
  \includegraphics[width=1\linewidth]{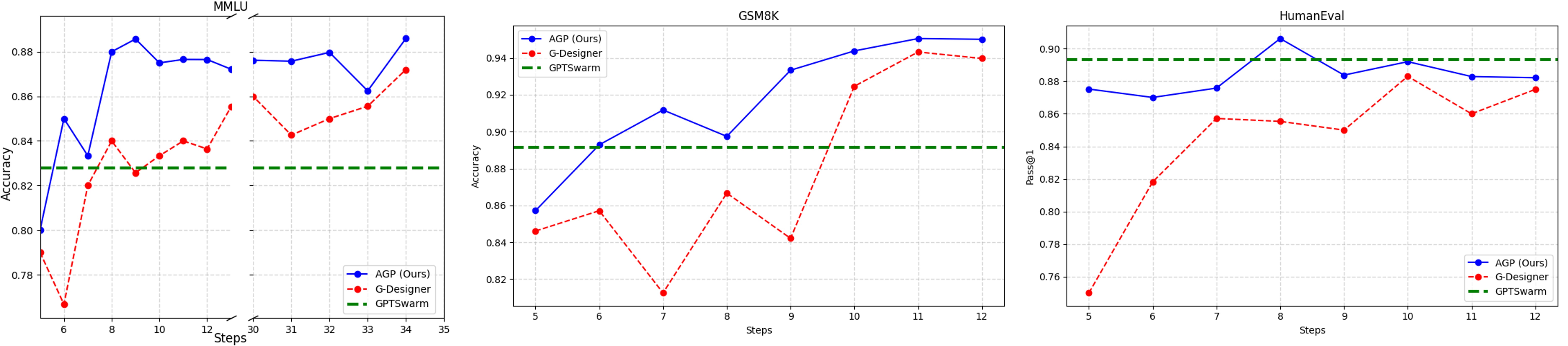}
  \caption{Under MMLU, GSM8k, and HummanEval benchmarks, the curves of the performance of \ourmethod and G-Designer as the number of training steps increases. Starting from the fifth step, there will be an evaluation after each batch is trained.}
   \label{fig:comparison}
   \vspace{4mm}
\end{figure*}

\subsection{Experiment Setup}

\paragraph{Tasks and Benchmarks.}
In order to make an objective evaluation, we evaluate \ourmethod on three categories of datasets corresponding to our training dataset: \textbf{(1) For general reasoning}, we use MMLU~\citep{mmlu}, which is a comprehensive multitask assessment featuring multiple-choice questions across various fields, including humanities, social sciences, hard sciences, and essential subjects like mathematics, US history, computer science, and law; \textbf{(2) For mathematical reasoning}, we choose GSM8K~\citep{arXiv2021_Verifier-Math}, a dataset of 8.5K high quality linguistically diverse grade school math word problems. We also select MultiArith~\citep{roy2016solving},  SVAMP~\citep{patel2021nlp}, and AQuA~\citep{ling2017program}, which are diverse in terms of language patterns and problem types, and is used to evaluate the model's ability to solve mathematical word problems; \textbf{(3) For code generation}, we evaluate on HumanEval~\citep{human-eval}, which consists of 164 original programming questions, assessing language understanding, algorithms and simple mathematics, as well as several types of questions for software interviews. 

\paragraph{Baselines.} 
To emphasize \ourmethod is generally superior to the existing works. We compare \ourmethod with a total of 12 agentic baselines of the two series: \textbf{(1) Single-agent Approaches}: Single-agent LLM, {CoT}~\citep{cot}, {ComplexCoT}~\citep{fu2022complexity}, and {Self-Consistency}~\citep{wang2023selfconsistency}; \textbf{(2)  Multi-agent Topologies}: MacNet~\citep{qian2024scaling}), AgentVerse~\citep{chen2023agentverse}, {MetaGPT}~\citep{meta-gpt}, {LLM-Debate}~\citep{arXiv2023_MultiAgent-Debate}, {LLM-Blender}~\citep{blender}, {DyLAN}~\citep{arXiv2023_Dynamic-LLM-Agent}, {GPTSwarm}~\citep{zhuge2024gptswarm}, and {G-Designer~\cite{zhang2024g}.

\paragraph{Implementation Details.} 
We access the GPT through the OpenAI API and mainly test on \llmname{gpt-4o-mini}. All models are accessed through APIs with the temperature set to 1. We also set a decision agent to aggregate the history of the dialogue and produce the final solution $a^{(K)}$ with $K = 3$ in all experiments. For all benchmarks, we use $Q \in\{100, 200\}$ queries for optimization. We conduct two evaluations of each baseline under the same benchmark and take the average as the final result. Here are some further details:

\begin{itemize} 
  \item \textbf{Mini‑batching.}\;
        10 supervision pairs $(\mathbf{A}^{\mathrm{gt}},\mathbf{y})$
        per batch keep GPU memory under~12 GB for $N_{\max}{=}16$.
  \item \textbf{Optimizer.}\;
        Adam, $\eta{=}10^{-3}$, $\beta_{1}{=}0.9$, $\beta_{2}{=}0.95$,
        weight decay $10^{-5}$, $10\sim20$ epochs.
  \item \textbf{Hyper‑parameters.}\;
        $\lambda_{\text{off}}{=}0.5,
         \lambda_s{=}0.1,
         \lambda_c{=}0.05,
         \beta{=}1.0$.
  \item \textbf{Class imbalance.}\;
        If fewer than $20\%$ of nodes are active in a batch,
        $\mathrm{BCE}$ switches to Focal Loss ($\gamma{=}2$) to reduce
        negative overwhelm.
\end{itemize}

\subsection{Quantitative Results}

We conduct extensive experiments in performance, training steps, and token consumption across six benchmarks to verify that \ourmethod is:

\paragraph{High‑performing.}
As summarized in \Cref{tab:rq1_performance}, \ourmethod delivers the strongest overall accuracy (\textbf{91.04}\%), surpassing the best static‑graph competitor G‑Designer by \textbf{+1.48} and the strongest single‑agent baseline (SC) by \textbf{+5.62}. It ranks \textbf{first on five of the six} task groups, MMLU, GSM8K, MultiArith, SVAMP, and AQuA, achieving per‑task gains ranging from \textbf{+2.58} (MultiArith) to \textbf{+9.84 } (MMLU). Even on \emph{HumanEval}, where the code‑centric MetaGPT specializes, \ourmethod attains nearly the best score (\textbf{90.62}\%), while using a single, unified topology learner rather than a domain‑specific workflow. The MultiArith result is particularly striking: \ourmethod solves \textbf{179 / 180} problems (\textbf{99.44}\%), indicating that the learned sparse graph preserves crucial numerical‑reasoning paths without excessive communication overhead.

\paragraph{Task‑adaptive.}~\label{sec:exp_task_ad}
In order to better explain the task-adaptiveness for multi-agent communication systems, we give the definition: A multi-agent communication system is \textbf{completely} task-adaptive if and only if the edge weights and the number of nodes in the communication topology can be changed simultaneously according to different tasks; and a multi-agent communication system is \textbf{partially} task-adaptive if and only if either the edge weights or the number of nodes in the communication topology can be changed according to different tasks.

The ``Mul.'' and ``Ada.'' columns reveal that methods able to \emph{both} coordinate multiple agents and tailor the topology per query enjoy the largest gains. \ourmethod is the only fully adaptive system (\fullsupp, \fullsupp) compared to all 12 baselines. Among all multi-agent communication systems, \ourmethod is the only one that improves or matches the best baseline across all domains, general knowledge (MMLU +9.84), mathematical reasoning (GSM8K +7.56, SVAMP +4.04, AQuA +9.78), and program synthesis (HumanEval +3.54  over single‑agent). Compared with DyLAN and GPTSwarm, which can adjust edges but not shrink the node set, \ourmethod gains \textbf{4.19} and \textbf{3.79} average accuracy, respectively, illustrating that joint edge–node pruning is key to per‑task efficiency. In short, the dual‑pruning strategy enables one model to generalize from commonsense QA to symbolic mathematics and near‑state‑of‑the‑art code generation with no domain‑specific tuning.

\paragraph{Token‑economical.}
\Cref{fig:token} contrasts each method’s accuracy (blue bars) with its average prompt‑token budget per query (pink bars).  Three trends emerge clearly.
(1) More tokens $\nRightarrow$ better accuracy for existing systems. LLM‑Debate~\citep{arXiv2023_MultiAgent-Debate}, DyLAN~\citep{arXiv2023_Dynamic-LLM-Agent}, and GPTSwarm~\citep{zhuge2024gptswarm} all follow a “bigger graph, longer prompt” recipe: on MMLU they spend between $1.2\!\times\!10^{6}$ and $2.6\!\times\!10^{6}$ tokens yet top out below $83\%$ accuracy, analogous patterns hold on GSM8K, SVAMP, and HumanEval.
(2) \ourmethod breaks the token–performance trade‑off.
With $2.5\!\times10^{5}$ tokens on MMLU, \textbf{90\% less} than GPTSwarm~\citep{zhuge2024gptswarm}, it raises accuracy from $82.8\%$ to \textbf{87.65\%} ($+4.85$).  Similar gains aear on GSM8K ($+1.04$ while cutting tokens by 65\%), SVAMP ($+5.28$, $-67\%$ tokens), and HumanEval ($+3.12$, $-22\%$ tokens).
(3) Compared to another adaptive graph (G‑Designer) \ourmethod is more frugal. G‑Designer~\citep{zhang2024g} already trims tokens aggressively, yet \ourmethod still reduces cost on every benchmark (\textit{e.g.}\ $0.48\!\times10^{7}$ vs.\ $0.55\!\times10^{7}$ on GSM8K) while adding $+1.04$ accuracy on average.
(4) Even when compared to MaAS~\citep{zhang2025multi}, a search algorithm that explores agent architecture distributions, our method achieves higher accuracy on GSM8K while reducing prompt-token costs by over half. Although it slightly lags behind MaAS in HumanEval due to MaAS’s retention of specific compile-and-test agents, our method still offers a better accuracy-cost trade-off overall.

These results confirm that jointly pruning \emph{edges and nodes} not only shortens dialogue transcripts but also steers attention to the most relevant agents. It shows that adaptive graph pruning can provide more accurate and economical solutions in complex settings without the iterative overhead of full architecture searches.

\begin{figure}[!t]
  \centering

  \begin{subfigure}[b]{0.49\linewidth}
    \includegraphics[width=\linewidth]{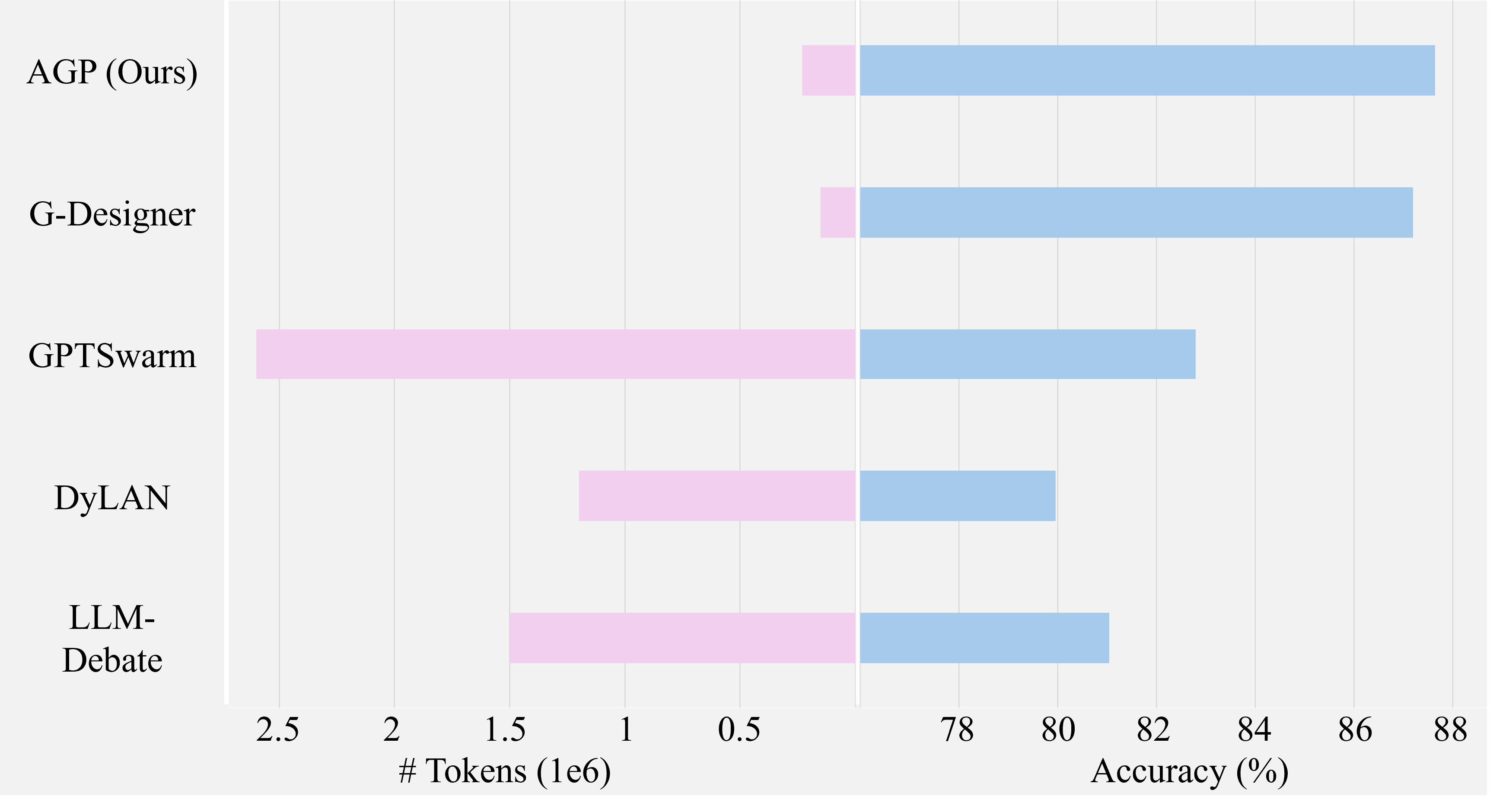}
    \caption{MMLU}
    \label{fig:MMLU}
  \end{subfigure}
  \hspace{0.1mm}
  \begin{subfigure}[b]{0.49\linewidth}
    \includegraphics[width=\linewidth]{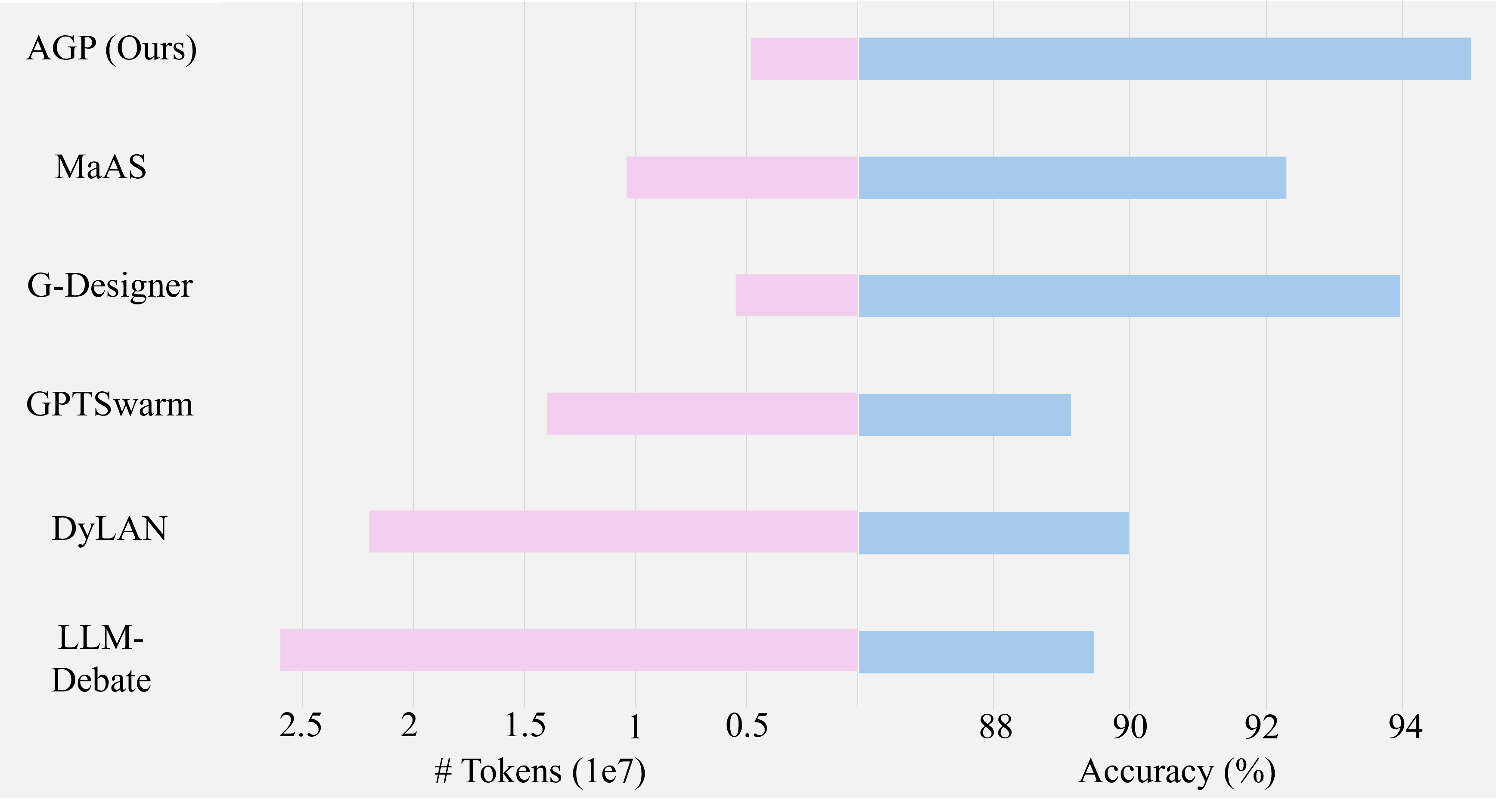}
    \caption{GSM8K}
    \label{fig:GSM8k}
  \end{subfigure}

  \vspace{4mm}

  \begin{subfigure}[b]{0.49\linewidth}
    \includegraphics[width=\linewidth]{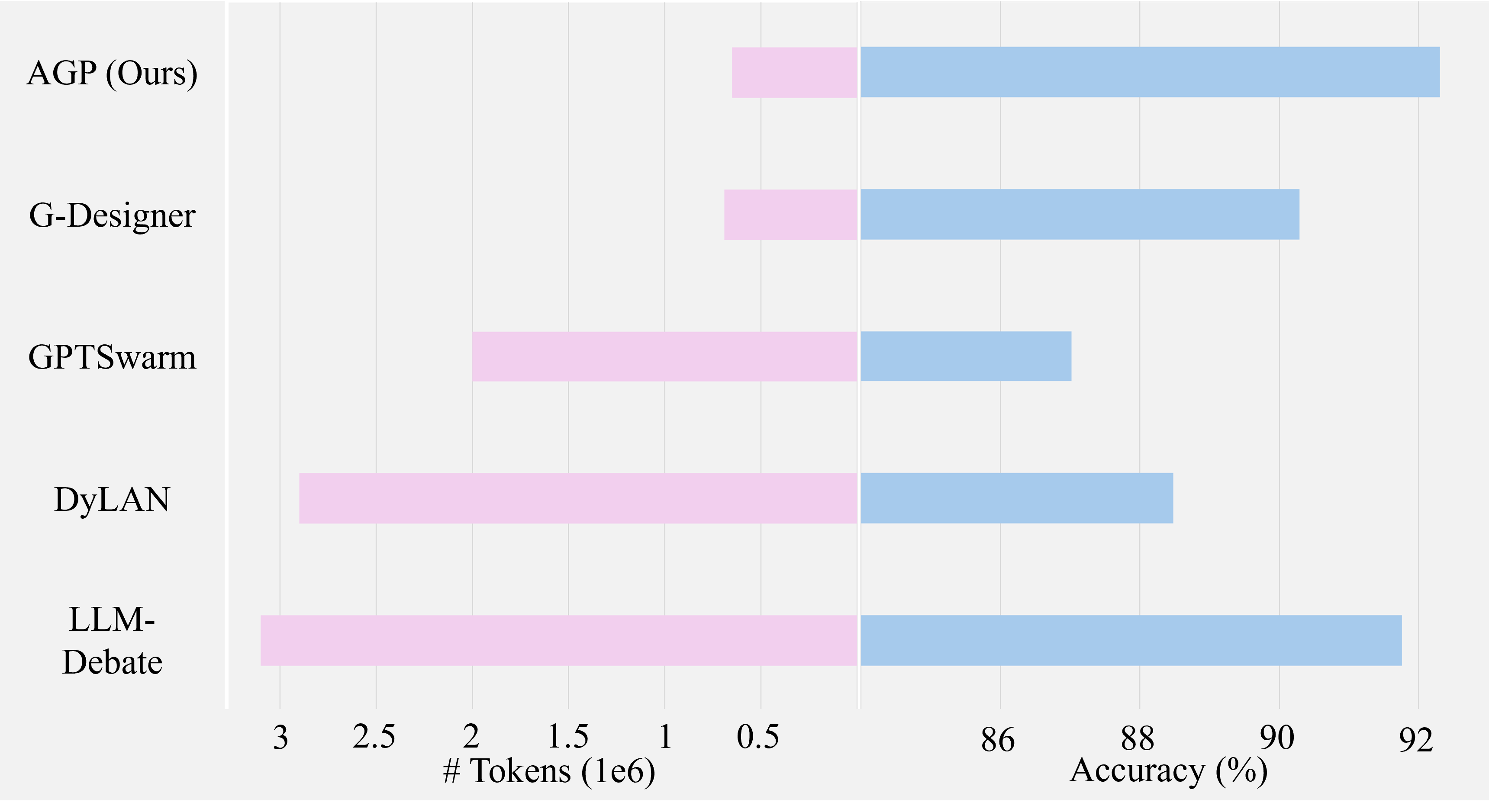}
    \caption{SVAMP}
    \label{fig:SVAMP}
  \end{subfigure}
  \hspace{0.1mm}
  \begin{subfigure}[b]{0.49\linewidth}
    \includegraphics[width=\linewidth]{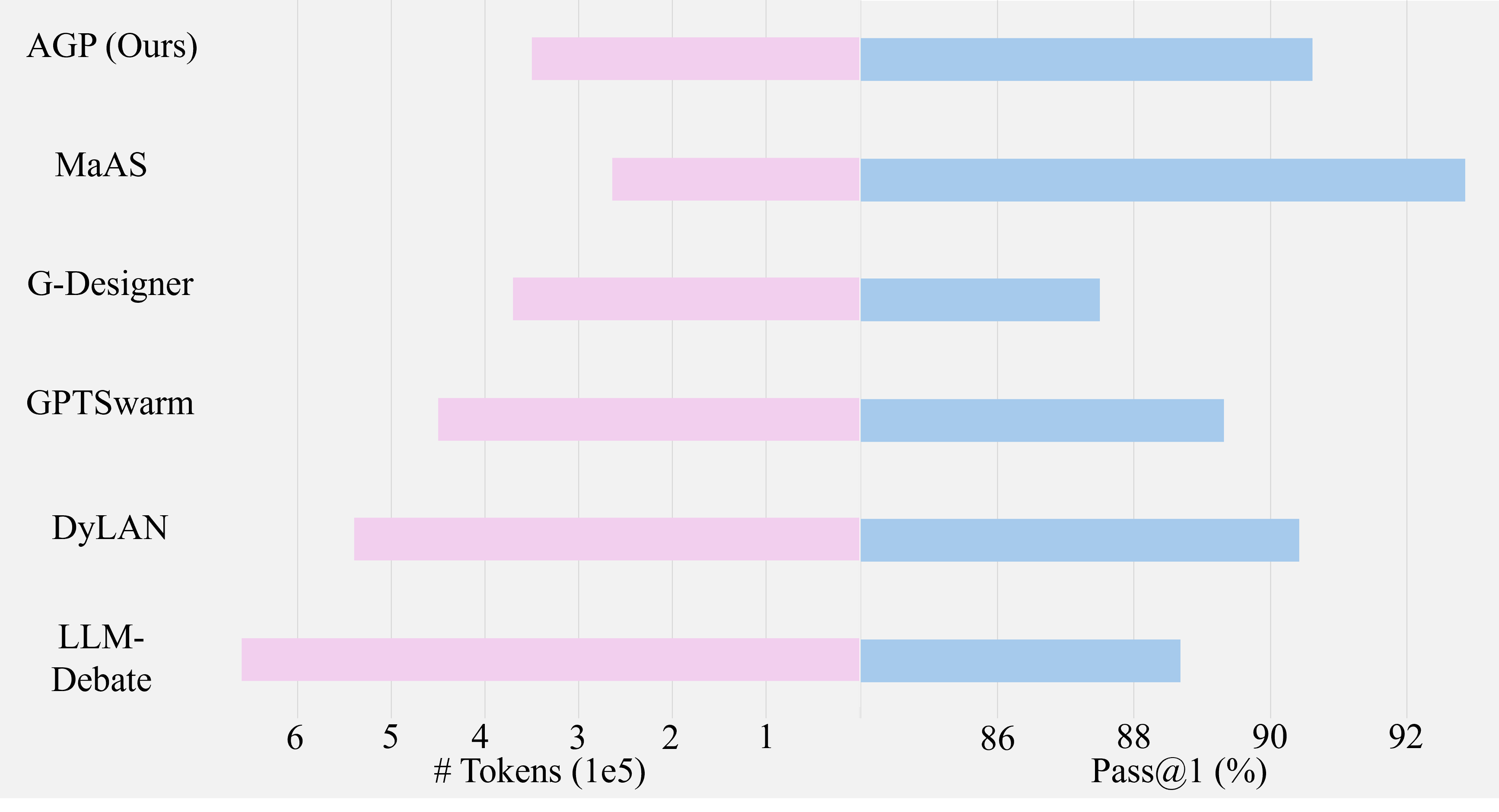}
    \caption{HumanEval}
    \label{fig:HumanEval}
  \end{subfigure}
  \vspace{5mm}
  \caption{Visualization of the performance and the number of prompt tokens of different multi-agent communication works across MMLU, GSM8K, SVAMP, and HumanEval benchmarks.}
  \label{fig:token}
  \vspace{4mm}
\end{figure}

\paragraph{Training‑efficient.}
\Cref{fig:comparison} plots accuracy (or Pass\@1) against training steps for \ourmethod, G‑Designer, and the fixed‑graph baseline GPTSwarm. On \textbf{MMLU} (left), \ourmethod climbs from 80\% to \textbf{88\%} in only 10 updates, then oscillates narrowly around that peak. G‑Designer~\citep{zhang2024g} starts lower (76\%), reaches its maximum of 87\% at step 34, but never matches our best score. We choose GPTSwarm~\cite{zhuge2024gptswarm} as our baseline, given its best score of 83\%. For \textbf{GSM8K} (center), both adaptive methods converge by step 12, but \ourmethod consistently outperforms G‑Designer~\citep{zhang2024g} at every checkpoint and reaches the 90\% barrier two steps earlier. A similar pattern appears on \textbf{HumanEval} (right): our curve remains above G‑Designer throughout training and surpasses the GPTSwarm~\citep{zhuge2024gptswarm} line after the very first evaluation. Taken together, the three curves show that \ourmethod attains higher final accuracy and baseline‑beating performance in fewer than ten optimization steps, evidencing markedly better sample‑ and compute‑efficiency during training.

\subsection{Case Study}

A natural question is whether \emph{human‑made} topologies, those that wire together exactly the agents a designer deems relevant, must always be the optimal choice.  To probe this, we inspect three representative tasks and compare the ``intuitive'' hand‑crafted profiles with the structures selected by \ourmethod. The outcomes fall into three categories, illustrated in~\Cref{fig:case_study}. More cases are shown in Appendix A.6~\cite{li2025adaptive}.

\paragraph{Case A (fully-intuitive).}
A primary‑school arithmetic word problem is solved with a minimal two‑agent chain: \emph{Algorithm Designer} produces the reasoning steps and \emph{Programming Expert} executes them.  The graph coincides with what a human would have drafted, confirming that \ourmethod{} does not over‑engineer simple tasks.

\paragraph{Case B (partially-intuitive).}
A simple code development task, \texttt{strlen(string)} only requires the core pair of agents, yet the initial pool also contains \emph{Bug Fixer} and \emph{Test Analyst}. \ourmethod{} prunes these extra nodes, recovering a subset of the human design and achieving the best score with fewer messages than any full four‑agent baseline.

\paragraph{Case C (counter‑intuitive).} A social‑and‑economic question appears unrelated to medicine, but the model retains both a \emph{Doctor} and the \emph{Programming Expert} alongside a \emph{Statistician}. Empirically, this three‑node graph outperforms all hand‑picked combinations, suggesting that seemingly ``irrelevant'' roles can inject orthogonal knowledge or critique that boosts final accuracy.

Together, these examples demonstrate that intuitive human layouts are \emph{not} always optimal. \ourmethod{} can (1) reproduce them when they suffice, (2) pare them down when redundancy exists, and (3) augment them with non‑obvious expertise when the task demands it, underscoring the framework’s strong task adaptivity and its ability to uncover useful but non‑trivial agent interactions.

\begin{figure}[!t]
  \centering
  \includegraphics[width=1\linewidth]{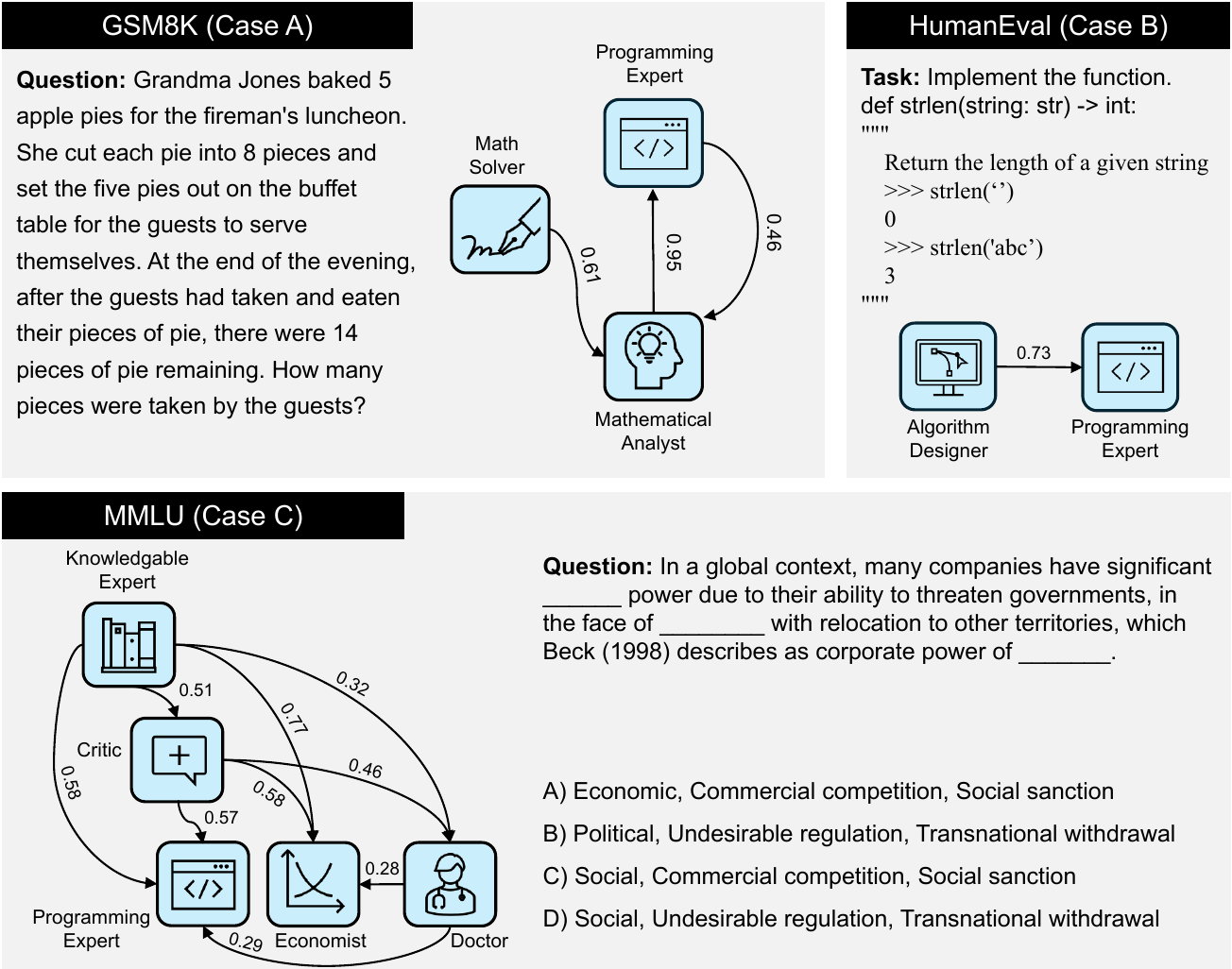}
  \caption{Case study of the communication topologies designed by \ourmethod on GSM8K, HumanEval, and MMLU benchmarks.}
   \label{fig:case_study}
   \vspace{5mm}
\end{figure}

\subsection{Ablation Study}

As shown in~\Cref{tab:variant_performance}, we disentangle the impact of the two pruning branches. When \emph{both} branches are enabled, \ourmethod{} sets the state of the art across three representative axes: general knowledge (MMLU~\textbf{87.65}), grade‑school mathematics (GSM8K~\textbf{95.01}), and function‑writing code synthesis (HumanEval~\textbf{90.62}).  Removing \textbf{soft‑pruning} but still letting the model decide which agents to keep leads to the steepest decline: {-4.16} on MMLU, {-4.88} on GSM8K, and {-4.99} on HumanEval.  The reason is that edge weights act as a continuous ``bandwidth throttle''; without them, every retained pair either exchanges an uncontrolled flood of messages or is shut off completely, so the system oscillates between information overload and starvation.
Conversely, disabling \textbf{hard‑pruning} while leaving edge weighting intact costs less accuracy, {-2.45}, {-3.45}, and {-2.21} on the same three tasks, because soft‑pruning can still attenuate many noisy links. Yet idle agents continue to produce prompts, wasting tokens and occasionally injecting spurious reasoning paths that drag down precision.  In addition, we observe a 22–30\% increase in prompt length for the ``w/o Hard'' variant, confirming the token penalty of carrying dead weight.
Taken together, the ablation shows a clear division of labor: \textbf{soft‑pruning} governs \emph{how loudly} surviving agents speak, whereas \textbf{hard‑pruning} decides \emph{who gets a voice} in the first place.  Only when these two levers operate in concert does the topology become both lean and expressive, unlocking the full benefit of task‑adaptive communication.


\begin{table}[!h]
    \centering
    \vspace{2mm}
    \caption{Ablation study of \ourmethod{} on MMLU, GSM8K, and HumanEval benchmarks.}
    \label{tab:variant_performance}
    \setlength{\tabcolsep}{6pt}
    \renewcommand\arraystretch{1.15}
    \rowcolors{2}{gray!10}{white}
    \resizebox{\linewidth}{!}{%
      \begin{tabular}{lccc}
        \toprule
        \rowcolor{gray!25}               
        \textbf{Variant} & \textbf{MMLU} & \textbf{GSM8K} & \textbf{HumanEval} \\
        \midrule
        Vanilla \ourmethod{}       & \textbf{87.65} & \textbf{95.01} & \textbf{90.62} \\
        \textit{w/o} Soft‑pruning  & 83.49 & 90.13 & 85.63 \\
        \textit{w/o} Hard‑pruning  & 85.20 & 91.56 & 88.41 \\
        \bottomrule
      \end{tabular}
    }
\end{table}

\section{Limitations}\label{sec:limitations}
We benchmark \ourmethod{} with a fixed \llmname{gpt-4o-mini} backend in the main paper and \llmname{gpt-3.5-turbo} in Appendix A.3 \cite{li2025adaptive}. Whether the dual-pruning policy transfers across other LLM families or model scales remains open. Aside from the node- and edge-analysis and weight in Appendix A.4 A.5~\cite{li2025adaptive}, richer quantitative analyses of efficiency are still needed. So far, all results are text-only, applying the adaptive graph to multimodal, temporally extended, embodied-agent tasks (vision–language, tool use, robotics simulations with sensor/action streams) could expose asymmetric bandwidth and other challenges.

\section{Conclusion}\label{sec:conclusion}
\ourmethod{} jointly soft-prunes edges and hard-prunes nodes to yield input-adaptive, variable-size communication graphs, a two-stage process, collecting task-optimized graphs, then training a dual-pruning GNN, lets a single forward pass produce a cost/accuracy-balanced topology. Across six benchmarks spanning knowledge QA, math reasoning, and code synthesis, it attains the best average score (\textbf{91.04}\%) while cutting prompt tokens by up to \textbf{90\%} and surpassing strong baselines in fewer than ten optimization steps. Once trained, \ourmethod{} removes exhaustive graph search and manual workflow design, acting as a plug-and-play module for heterogeneous agent pools and a practical path toward resource-aware multi-agent LLM ecosystems.



\section{Acknowledgement}

This work was supported by the Zhejiang Provincial Natural Science Foundation of China (LZ24F030005, LD24F020016), the Scientific Research Foundation of Sichuan Provincial Department of Science and Technology, China (2024YFHZ0001), and the Research Fund for International Scientists of National Natural Science Foundation of China (72350710798).

\appendix
\renewcommand\thefigure{\Alph{section}\arabic{figure}}
\renewcommand\thetable{\Alph{section}\arabic{table}}
\setcounter{figure}{0}
\setcounter{table}{0}
\titlespacing*{\paragraph}{0pt}{1ex plus .2ex minus .2ex}{1em}

\renewcommand{\thesection}{A}

\section{Appendix}

\paragraph{The appendix is organized as follows:}
\begin{itemize}
    \item \textbf{Motivation} summarizes the motivation details of our work. We analyze existing methods and problems in the field of multi-agent communication systems, explaining our mental journey on why we decided to do \ourmethod (\Cref{sec:supp_motivation}).
    \item \textbf{Notations} provide detailed notations use in our main text (\Cref{sec:supp_notation}).
    \item \textbf{Performance on \llmname{gpt-3.5-turbo}} provides the performance of \ourmethod with the base model \llmname{gpt-3.5-turbo} under MMLU and GSM8K to test its ability on general reasoning and mathematical reasoning. It also proves the generalizability of \ourmethod to other LLM architectures in some way(\Cref{sec:gpt3.5}).
    \item \textbf{Analysis of the number of nodes and edges} provides a detailed analysis of the changes in the number of nodes and edges after pruning. Such an analysis would be beneficial for understanding the diverse roles and contributions of various agents within the system(\Cref{sec:supp_analysis}).
    \item \textbf{Discussion on weights} provides a discussion on the influence of \(\beta\) on the performance of \ourmethod in the loss formula provided in \Cref{sec:tot-obj}(\Cref{sec:supp_discussion}).
    \item \textbf{Further Case Study} provides a more detailed account of the research we discovered and conducted on the three types of cases (fully-intuitive, partially-intuitive, and counter-intuitive) based on the article (\Cref{sec:supp_case}).
    \item \textbf{Data Statistics} concludes the statistics of the training dataset and evaluation datasets we use in the experiments (\Cref{sec:supp_data_stat}).
    \item \textbf{Agent Profile Details} introduces the agents in \ourmethod and the corresponding prompts (\Cref{sec:profiles}).
\end{itemize}

\subsection{Motivation}\label{sec:supp_motivation}

Selecting a task‑appropriate communication graph is the key obstacle to scaling LLM‑agent systems.  Current approaches fall into two camps. \emph{Hand‑crafted} topologies: chains \citep{cot,meta-gpt}, trees \citep{tot,autogen}, stars \citep{autogen}, complete or random graphs \citep{qian2024scaling}, encode strong human priors that transfer poorly across domains and require manual retuning whenever the task distribution changes \citep{zhuge2024gptswarm,zhang2024g}. \emph{Edge‑learning} methods relax this rigidity by optimizing link weights inside a fixed agent pool \citep{zhuge2024gptswarm,zhang2024g}, \textit{i.e.}, perform \textit{soft‑pruning}.  Yet they implicitly assume that keeping \emph{all} agents is always harmless.  Our reproductions contradict this: on GSM8K, HumanEval, and MMLU, accuracy peaks only for a narrow team size, adding seemingly ``irrelevant'' agents hurts both accuracy and token cost. Hence, \emph{who} participates (a \textit{hard‑pruning} decision) is as important as \emph{how strongly} they communicate.
These observations raise two research questions:

\begin{itemize}
\item \textbf{Q1.} Can we design a topology learner that \emph{jointly} selects the optimal agent subset \textit{and} their edge weights directly from task feedback, eliminating manual role curation?
\item \textbf{Q2.} Are human‑intuitive agent combinations always optimal, or do counter‑intuitive mix-agents that seem unrelated to the task sometimes yield superior performance?
\end{itemize}

\ourmethod{} addresses both questions by unifying hard‑ and soft‑pruning in a single end‑to‑end framework. The case study in the main text presents concrete case studies that illustrate how the learned graphs can reproduce, refine, or even overturn human intuition.

\subsection{Notations}~\label{sec:supp_notation}

To keep the exposition compact, we use a small set of symbols that re‑appear throughout Stage I (graph collection), Stage II (dual pruning), and the experiments. \Cref{tab:our_notations} groups them by function. The top block defines the \emph{search space}: a max‑complete graph $K_{N_{\max}}$ with anchored agents $\mathcal V$ and its sub‑graph family $\mathfrak G$.  The middle block covers the two \textbf{degrees of freedom} that \ourmethod learns—edge weights $\mathbf W$ for \emph{soft‑pruning} and the node mask $\mathbf m$ for \emph{hard‑pruning}.  Together they yield the task‑adaptive topology $\mathcal G_{\mathrm{com}}$.  The last block lists the optimization components: utility $U$, cost $C$, the trade‑off coefficient $\lambda_c$, supervision pairs $(\mathbf A^{\mathrm{gt}},\mathbf y)$ obtained in Stage I, and the node:edge loss terms that form the total objective.  These notations are consistent across all formulae, algorithms, and tables in the rest of the paper, so readers can refer back to \Cref{tab:our_notations} whenever an unfamiliar symbol appears.

\subsection{Performance on \llmname{gpt-3.5-turbo}}\label{sec:gpt3.5}

In addition to \llmname{gpt-4o-mini}, here we conduct experiments on \llmname{gpt-3.5-turbo} under the same benchmarks. As shown in~\Cref{tab:gpt3.5}, it achieves an accuracy of 76.61\% under MMLU, 87.27\% under GSM8K, and an average accuracy of 81.94\%, which is also 0.29\% higher than the single-agent with \llmname{gpt-4}. It confirms that \ourmethod can indeed improve the performance. Besides, it provides evidence for the generalizability of \ourmethod to other LLM architectures.

\begin{table}[!h]
  \centering
  \caption{Performance comparison with \llmname{gpt-4} and \ourmethod under \llmname{gpt-3.5-turbo}. We \textbf{bold} best results and \underline{underline} runner‑ups. ``Mul.'' and ``Ada.'' indicate multi‑agent support and task adaptivity, respectively. \nosupp, \partsupp, and \fullsupp denote no, partial, and full support.}
  \label{tab:gpt3.5}

  \setlength{\tabcolsep}{5.3pt}
  \renewcommand\arraystretch{1.15}

  \rowcolors{2}{gray!10}{white}

  \footnotesize
  \resizebox{\linewidth}{!}{%
  \begin{tabular}{lcclll}   
    \toprule
    \rowcolor{gray!25}
    \textbf{Method} & \textbf{Mul.} & \textbf{Ada.} & \textbf{MMLU} & \textbf{GSM8K} & \textbf{Avg.} \\
    \midrule
    Single‑agent(\llmname{gpt-4}) & \nosupp  & \nosupp  & \textbf{77.90} & \underline{85.40} & \underline{81.65}\\
    \midrule
    \ourmethod(\llmname{gpt-3.5-turbo}) & \fullsupp & \fullsupp & \underline{76.61}\blue{1.29} & \textbf{87.27}\red{1.87} & \textbf{81.94}\red{0.29}\\
    \bottomrule
  \end{tabular}}
\end{table}

\begin{table*}[!t]
\centering
\caption{Symbols used in \ourmethod.}
\label{tab:our_notations}
\renewcommand\arraystretch{1.1}
\setlength{\tabcolsep}{5pt}
\rowcolors{2}{gray!10}{white}
\footnotesize
\begin{tabular}{lp{11.7cm}}
\rowcolor{gray!10}
\toprule
\textbf{Notation} & \textbf{Meaning in the context of Adaptive Graph Pruning (\ourmethod)} \\
\midrule
$N_{\max}$                 & Size of the heterogeneous agent pool; also the order of the max‑complete graph $K_{N_{\max}}$. \\
$\mathcal{V}\!=\!\{v_i\}$   & Anchored vertices (agents). Each $v_i=\langle\mathrm{LM}_i,r_i,s_i,\phi_i\rangle$. \\
$\mathbf{A}\in\{0,1\}^{N_{\max}\times N_{\max}}$ & Binary adjacency of $K_{N_{\max}}$; $A_{ij}=1$ iff $(v_i{\rightarrow}v_j)$ is allowed. \\
$\mathbf{W}\in[0,1]^{N_{\max}\times N_{\max}}$   & Learned edge‑weight matrix after \textbf{soft‑pruning}. \\
$\mathbf{m}\in\{0,1\}^{N_{\max}}$               & Node‑retention mask from \textbf{hard‑pruning}; $m_i=1$ keeps agent $v_i$. \\
$\mathcal{G}=(\mathcal{V},\mathbf{A})$           & Max‑complete graph that spans all agents. \\
$\mathcal{G}[\mathbf{m}]$                       & Node‑induced sub‑graph obtained by hard‑pruning with mask $\mathbf{m}$. \\
$\mathcal{G}_{\mathrm{com}}=(\mathcal{G}[\mathbf{m}],\mathbf{W})$ & Final task‑adaptive communication topology produced at inference. \\
$\mathfrak{G}$                                   & Space of all sub‑graphs of $K_{N_{\max}}$ (\textit{i.e.}\ all $\mathcal{G}_{\mathrm{com}}$ candidates). \\
$\mathcal{Q}$                                   & Incoming query (task instance). \\
$a^{(K)}$                                       & Team output after $K$ communication rounds. \\
$U(\mathbf{A}\!\mid\!\mathcal{Q})$              & Utility of topology $\mathbf{A}$ on query $\mathcal{Q}$ (e.g.\ accuracy). \\
$C(\mathbf{A})$                                 & Communication cost (token count) incurred by $\mathbf{A}$. \\
$\lambda_c$                                     & Trade‑off weight between utility and cost. \\
$(\mathbf{A}^{\mathrm{gt}},\mathbf{y})$          & Ground‑truth edge labels and node mask mined in Stage I for supervision. \\
$\mathcal{L}_{\text{edge}},\mathcal{L}_{\text{node}}$ & Loss terms for soft‑ and hard‑pruning. \\
$\mathcal{L}_{\text{total}}$                    & Joint training objective $\mathcal{L}_{\text{edge}} + \beta\mathcal{L}_{\text{node}}$. \\
$\tau$                                          & Temperature of the Gumbel–Sigmoid continuous–discrete bridge. \\
$B$                                             & Budget size of the sampled graph pool in Stage I. \\
\bottomrule
\end{tabular}
\end{table*}

\subsection{Analysis of the number of nodes and edges}\label{sec:supp_analysis}

We set a max-complete graph with 9 agents, respectively studying situations where a large (general reasoning) or small (mathematical reasoning and code generation) number of agents is needed, and performing Gaussian distribution fitting on data with node quantity as the x-axis and topology quantity as the y-axis. For the former case, we get \(A = 46.834, \mu = 8.027, \sigma = 1.310\), which is slightly left-skewed. For the latter case, we get \(A = 64.868, \mu = 3.594, \sigma = -0.607\), which is nearly standard(\(A, \mu, \sigma\) respectively represent the maximum value, mean, and standard deviation of the function). This result can further prove that \ourmethod can provide more economical and task-adaptive solutions.

\subsection{Discussion on weights}\label{sec:supp_discussion}

Our motivation stems from the fact that we found that agent quantity had a significant impact on the final performance, and finding a high-performance method for jointly optimizing soft- and hard-pruning is crucial. The addition of hard-pruning can be regarded as changing the \(\beta\) (node:edge) in the loss formula in \Cref{sec:tot-obj} from 0 to 1, so we think giving more weight to nodes over edges is intuitive. Experiments from \Cref{sec:exp}have also proved that \ourmethod does bring performance improvements and other advantages.

In further experiments, we set \(\beta\) (node:edge) to 0.75, 1, and 1.333, respectively. The results show that under the same number of training steps, more edge weights converge more quickly (batch size = 10, steps = 20, learning rate = 1e-3). The final converged loss = 0.723 when \(\beta = 1\), and is 24.2\% higher than that when \(\beta = 0.75\) (loss = 0.582). When \(\beta = 1.333\), the loss curve oscillates and does not converge.

\subsection{Further Case Study}~\label{sec:supp_case}

To further discuss whether \emph{human‑made} topologies—those that wire together exactly the agents a designer deems relevant, must always be the optimal choice, we conducted a further case study, which inspects three representative tasks (fully-intuitive, partially-intuitive, and counter‑intuitive) for each type of task in general reasoning, mathematical reasoning, and code generation by \ourmethod, illustrated in~\Cref{fig:further1}, ~\Cref{fig:further2}, and~\Cref{fig:further3}.

\paragraph{Case A (fully-intuitive).}
In \textbf{Case A}, the communication topologies \ourmethod generated coincide with what a human would have drafted, confirming that \ourmethod{} does not over‑engineer simple tasks.

\paragraph{Case B (partially-intuitive).}
In \textbf{Case B}, \ourmethod{} prunes extra nodes for those easy tasks, recovering a subset of the human design and achieving the best score with fewer messages than any full four‑agent baseline.

\paragraph{Case C (counter‑intuitive).}
\textbf{Case C} illustrates that \ourmethod will sometimes pick seemingly ``irrelevant'' roles for communication topologies, which can inject orthogonal knowledge or critique that boosts final accuracy. This further demonstrates the high performance of \ourmethod and the irrationality of the human-designed communication topologies.

The presented examples collectively illustrate that intuitive human-designed communication topologies are not always optimal. \ourmethod can \textbf{(1)} replicate such topologies when they are sufficient, \textbf{(2)} streamline them when redundancy is identified, and \textbf{(3)} enhance them by introducing agents that seem counterintuitive but might be related to the task. This underscores the \ourmethod’s robust adaptability to various tasks and its capacity to reveal valuable yet non-obvious agent interactions.

\begin{figure}[!h]
  \centering
  \includegraphics[width=1\linewidth]{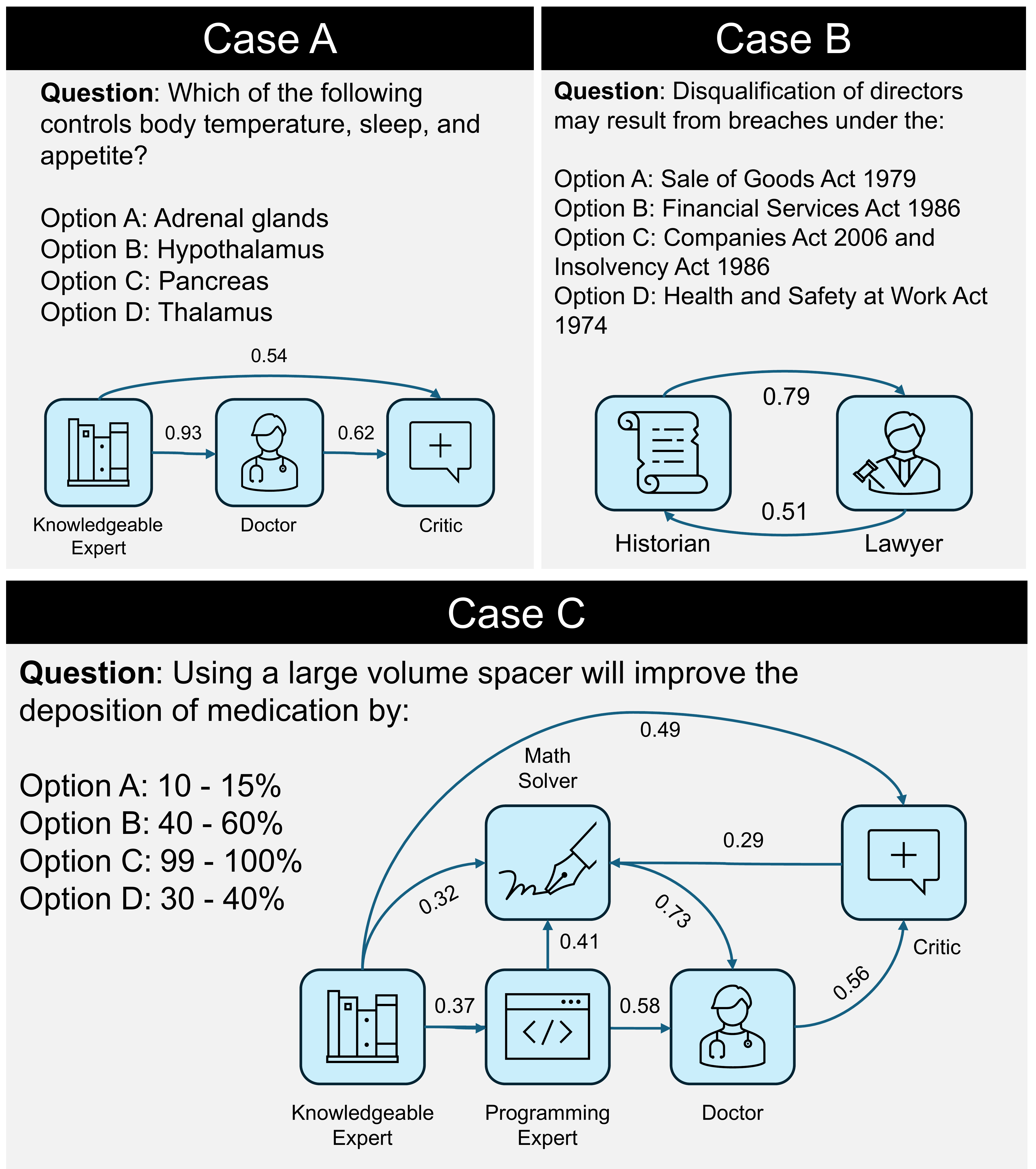}
  \caption{Further case study for general reasoning tasks, grouped by how the optimal agent set compares to a hand‑designed ``intuitive'' roster: \textbf{fully‑intuitive}, \textbf{partially‑intuitive}, and \textbf{counter‑intuitive}.}
   \label{fig:further1}
   \vspace{5mm}
\end{figure}

\begin{figure}[!h]
  \centering
  \includegraphics[width=1\linewidth]{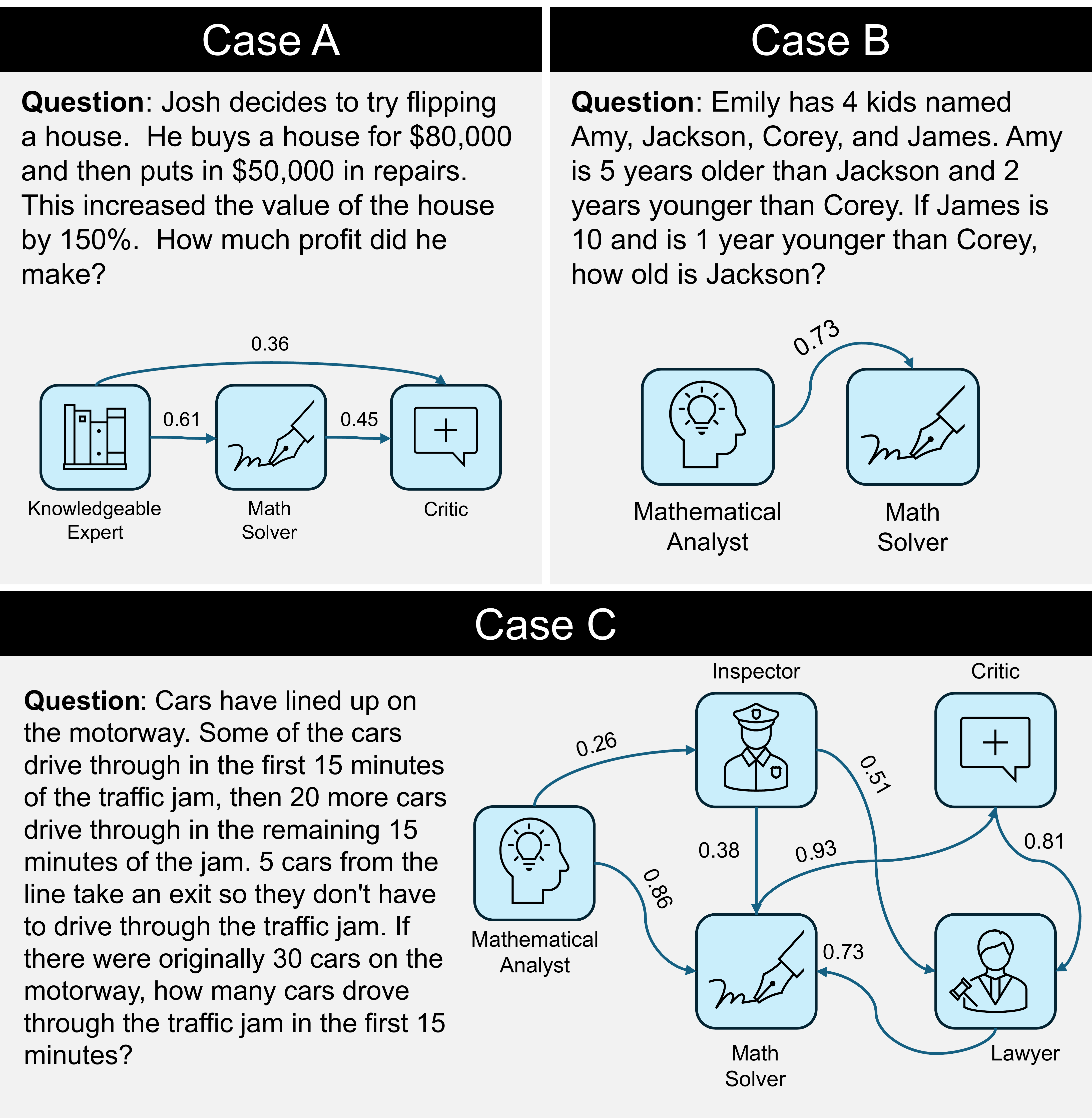}
  \caption{Further case study for mathematical reasoning tasks, grouped by how the optimal agent set compares to a hand‑designed ``intuitive'' roster: \textbf{fully‑intuitive}, \textbf{partially‑intuitive}, and \textbf{counter‑intuitive}.}
   \label{fig:further2}
   \vspace{5mm}
\end{figure}

\begin{figure}[!h]
  \centering
  \includegraphics[width=1\linewidth]{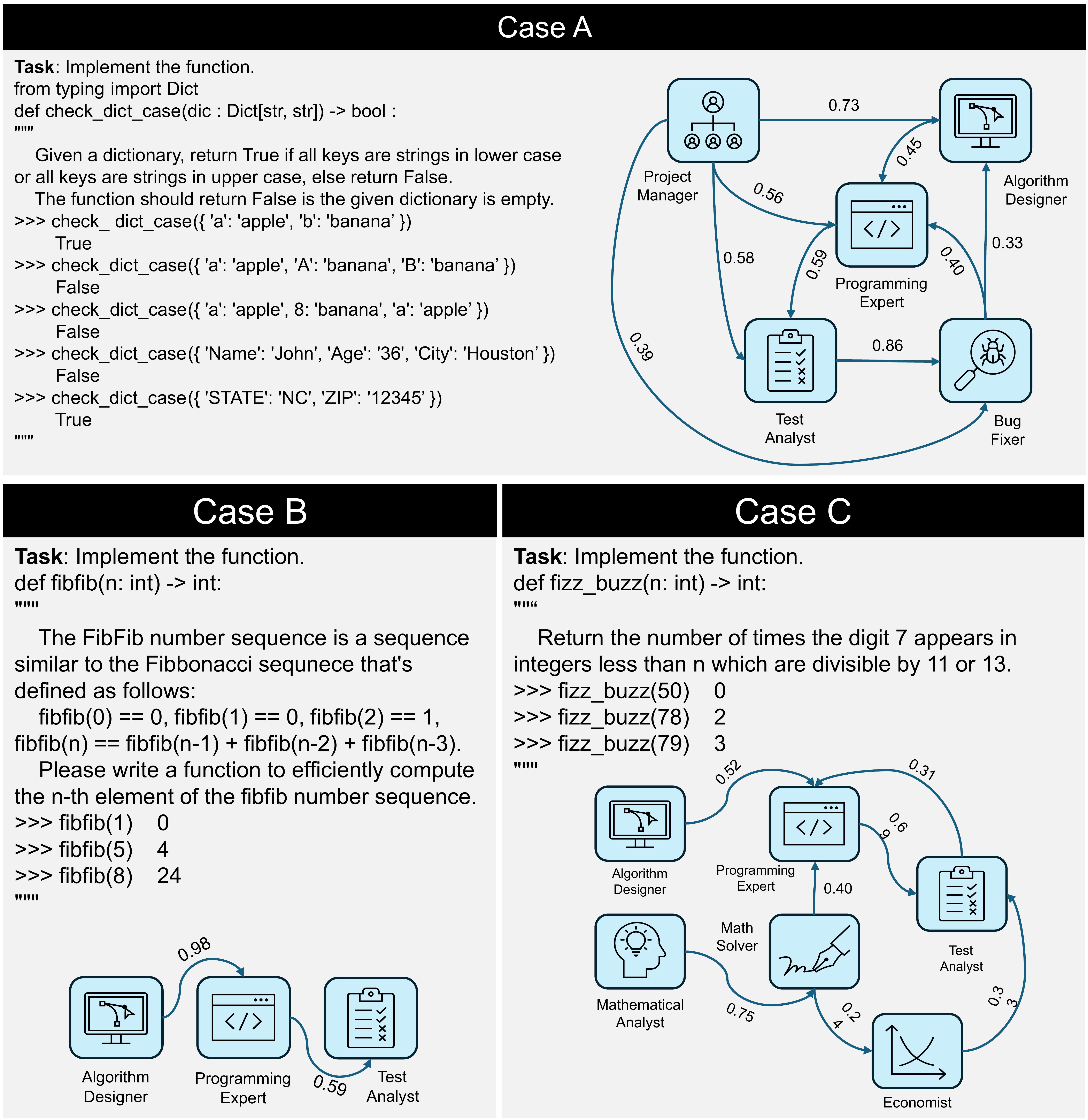}
  \caption{Further case study for code generation tasks, grouped by how the optimal agent set compares to a hand‑designed ``intuitive'' roster: \textbf{fully‑intuitive}, \textbf{partially‑intuitive}, and \textbf{counter‑intuitive}.}
   \label{fig:further3}
   \vspace{5mm}
\end{figure}

\subsection{Data Statistics}~\label{sec:supp_data_stat}

\begin{figure}[!h]
  \centering
  \includegraphics[width=1\linewidth]{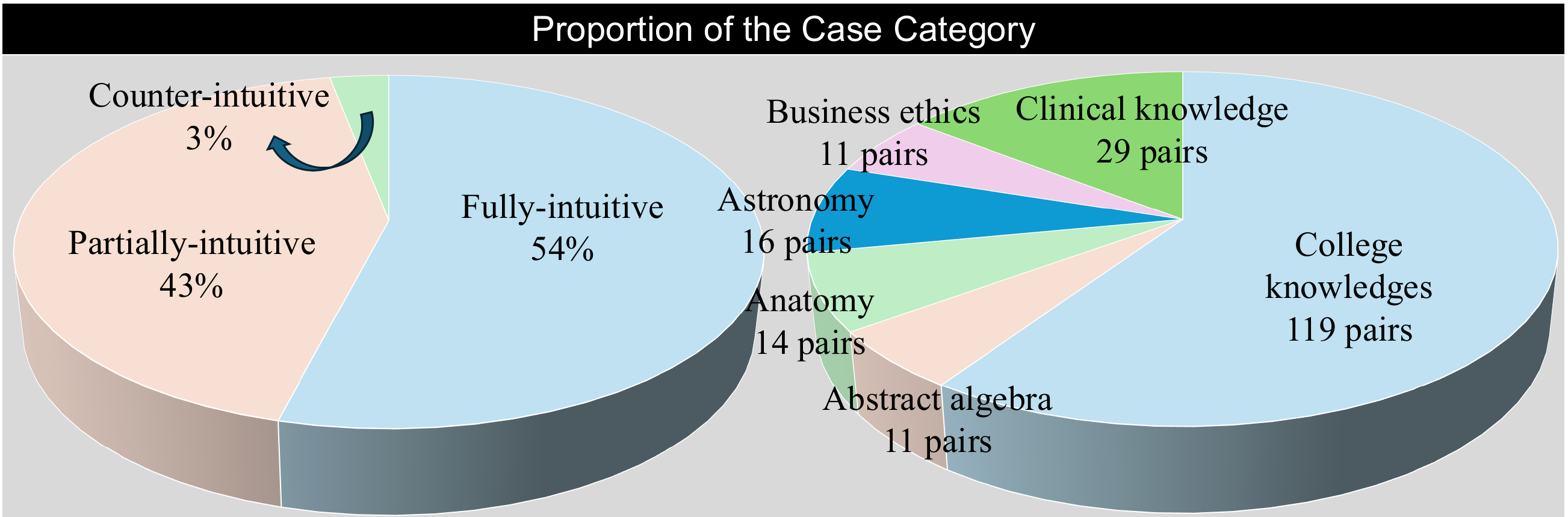}
  \caption{Distribution of the supervision pairs collected in Stage I, grouped by how the optimal agent set compares to a hand‑designed ``intuitive'' roster: \textbf{fully‑intuitive}, \textbf{partially‑intuitive}, and \textbf{counter‑intuitive}.}
   \label{fig:case_propotion}
   \vspace{5mm}
\end{figure}

\vspace{-10mm}
\subsubsection{Training‑set Statistics}\label{sec:sup_traindata_stats}
\vspace{5mm}
\begin{table}[h]
    \centering
    \caption{Training-set descriptions and statistics.}
    \label{tab:training_dataset_stats}
    \resizebox{\linewidth}{!}{%
        \begin{tabular}{llccc}
            \toprule
            \rowcolor{gray!10}
            \textbf{Category} & \textbf{Subtasks} & \textbf{Answer Type} & \textbf{Metric} & \textbf{\#Test} \\
            \midrule
            General reasoning & \makecell[l]{College knowledges, \\ Abstract algebra, \\ Anatomy, \\ Astronomy, \\ Business ethics, \\ Clinical knowledge} & Multi-choice & Acc. & 200 \\
            \midrule
            Math reasoning & \makecell[l]{Elementary school math problems, \\Math word problems} & Number & Acc. & 100 \\
            \midrule
            Code generation & Complete code to passe the tests & Code & Pass@1 & 160 \\
            \bottomrule
        \end{tabular}
    }
\end{table}

We conclude the training-set statistics in~\Cref{tab:training_dataset_stats}, and we have analyzed the data we collected as shown in~\Cref{fig:case_propotion}.

\Cref{fig:case_propotion} summarizes how often the ground‑truth graphs collected in Stage I overlap with a hand‑crafted, “intuitive’’ agent layout.  Roughly \textbf{54\%} of the tasks fall into the \emph{fully‑intuitive} slice: the optimal graph matches exactly what a human would wire up.  A further \textbf{43\%} are \emph{partially‑intuitive}: the best solution is a \emph{subset} of the human design, redundant agents increase token cost without improving accuracy. The remaining \textbf{3\%} are \emph{counter‑intuitive}: top performance is achieved only when the graph retains agents that appear unrelated to the task. These outliers echo the logic of the movie \textbf{``12 Angry Men''}: diversity of expertise can surface orthogonal knowledge that lifts group reasoning.

The distribution highlights two insights. First, almost one half of real‑world tasks \emph{cannot} be solved optimally by a pre‑defined, human‑intuitive team, underscoring the need for automatic \textbf{hard‑pruning} to trim redundant roles. Second, the small but non‑negligible counter‑intuitive slice shows that occasionally adding a seemingly irrelevant agent is beneficial, an effect that our joint soft–/hard‑pruning framework can capture, whereas fixed or manually pruned systems would miss.

\vspace{-3mm}
\subsubsection{Benchmark Statistics}
We conclude the benchmark statistics in \Cref{tab:dataset_stats}
\begin{table}[h]
    \centering
    \caption{Benchmark descriptions and statistics.}
    \label{tab:dataset_stats}
    \resizebox{\linewidth}{!}{%
        \begin{tabular}{llcccl}
            \toprule
            \rowcolor{gray!10}
            \textbf{Category} & \textbf{Dataset} & \textbf{Answer Type} & \textbf{Metric} & \textbf{\#Test} & \textbf{License} \\
            \midrule
            \multirow{1}{*}{General reasoning} & MMLU & Multi-choice & Acc. & 153 & MIT License \\
            \midrule
            \multirow{4}{*}{Math reasoning} 
            & GSM8K & Number & Acc. & 1,319 & MIT License \\
            & MultiArith & Number & Acc. & 600 & Unspecified \\
            & SVAMP & Number & Acc. & 1,000 & MIT License \\
            & AQuA & Multi-choice & Acc. & 254 & Apache-2.0 \\
            \midrule
            \multirow{1}{*}{Code generation} & HumanEval & Code & Pass@1 & 164 & MIT License \\
            \bottomrule
        \end{tabular}
    }
\end{table}

\vspace{-5mm}
\subsection{Agent Profile Details}\label{sec:profiles}

\begin{table}[!h]
\centering
\caption{Full roster of agent profiles on $K_{15}$.}
\label{tab:agent_profiles}
\renewcommand\arraystretch{1.1}
\setlength{\tabcolsep}{4.2pt}
\rowcolors{2}{gray!10}{white}
\footnotesize
\begin{tabular}{lll}
\rowcolor{gray!10}
\toprule
\textbf{\#} & \textbf{Role} & \textbf{Primary Expertise / Duty} \\
\midrule
1  & Know\-ledgeable Expert & Suggest key entities for external search \\ 
2  & Critic                 & Point‑by‑point flaw inspection            \\
3  & Psychologist           & Provide psycho‑social advice              \\
4  & Historian              & Analyse past cultural \& political events \\
5  & Doctor                 & Recommend treatments and remedies        \\
6  & Lawyer                 & Legal and policy reasoning                \\
7  & Economist              & Macro‑/micro‑economic analysis            \\
8  & Project Manager        & High‑level code structure planning        \\
9  & Algorithm Designer     & Detailed algorithm design \& pseudocode   \\
10 & Test Analyst           & Generate edge‑case tests and critiques     \\
11 & Bug Fixer              & Produce corrected Python implementations   \\
12 & Math Solver            & Step‑by‑step symbolic math derivation      \\
13 & Mathematical Analyst   & Variable‑level proof and numeric check     \\
14 & Programming Expert     & End‑to‑end code authoring                  \\
15 & Inspector              & Cross‑check reasoning and code consistency\\
\bottomrule
\end{tabular}
\end{table}
\subsubsection {Heterogeneous Agent Pool}

\Cref{tab:agent_profiles} lists the fifteen roles that anchor the
\emph{max‑complete} graph $K_{15}$ used throughout Stage I.
Each role is assigned a concise expertise tag and a one‑line
Responsibility summary distilled from the full prompts in other supplementary materials.

\subsubsection{Coverage Statistics}

\begin{itemize}
\item \textbf{Domain breadth.}\;
      The pool spans \emph{five} knowledge clusters—
      humanities (Historian, Lawyer), social science (Economist,
      Psychologist), STEM (Math Solver, Mathematical Analyst,
      Algorithm Designer), software engineering
      (Project Manager, Programming Expert, Bug Fixer, Test Analyst,
      Inspector), and general reasoning/critique (Knowledgeable
      Expert, Critic). 
\item \textbf{Role specialization.}\;
      7 / 15 agents focus on \textbf{code and algorithm} tasks,
      4 / 15 on \textbf{factual or legal knowledge},
      2 / 15 on \textbf{mathematical deduction},
      and the remaining 2 provide \textbf{meta‑reasoning}
      (search guidance and critical review).
\item \textbf{Graph semantics.}\;
      During training the complete graph allows \emph{all} 105 directed
      edges to be considered; soft‑/hard‑pruning subsequently
      attenuates edge weights and removes nodes to yield task‑specific
      sub‑graphs.
\end{itemize}


\begin{figure*}[t]
\begin{tcolorbox}[colback=white, colframe=black, boxrule=0.5pt, sharp corners=all, title=\textbf{Prompt Format for a Single Agent of \ourmethod in mathematical Reasoning}]
\setlength{\parskip}{1ex}

\textbf{Input Format:}
\begin{itemize}
    \item \textbf{Task}: \textcolor{blue}{Given current task}
    \item \textbf{Profile}: \textcolor{orange}{The profile for the corresponding agent.}
    \item \textbf{Dialogue History}: The output of all agents from the beginning to the present.
\end{itemize}

\textbf{Output Format:}
\begin{itemize}
    \item \textbf{The Reasoning}: The thinking process regarding the problem and the previous conversation records.
    \item \textbf{Answer}: The answer in its own opinion.
\end{itemize}

\textbf{Prompts for Input:}
\begin{tcolorbox}[colback=gray!5, colframe=gray!50!black]
\begin{itemize}
    \item \textbf{System Prompt:} \begin{verbatim} 
<Profile>. And your task is to solve the question: <Task>. 
    \end{verbatim}
    \item \textbf{User Prompt:} \begin{verbatim}
At the same time, there are the following responses to the same
question for your reference: <Dialogue History>.
    \end{verbatim}
\end{itemize}
\end{tcolorbox}

\textbf{Agent Output:}
\begin{tcolorbox}[colback=gray!5, colframe=gray!50!black]
\begin{verbatim}
In my opinion, I think <The Reasoning>.
And my answer to the <Task> is <Answer>.
\end{verbatim}
\end{tcolorbox}

Each agent will update \texttt{<Dialogue History>} after executing

\textbf{Fields to be included in JSON:}
\begin{itemize}
    \item \texttt{id}, \texttt{role}, \texttt{Agent Output}
\end{itemize}

\textbf{Dialogue History Format:}
\begin{verbatim}
{
  "Dialogue History": [
    {
        "id": A8MK,
        "role": "Math Solver",
        "output": "In my opinion, I think <The Reasoning>. 
        And my answer to the <Task> is <Answer>."
    },
    {
        "id": X2LJ,
        "role": "Programming Expert",
        "output": "In my opinion, I think <The Reasoning>. 
        And my answer to the <Task> is <Answer>."
    }
  ]
}
\end{verbatim}

\end{tcolorbox}
\label{fig:prompt}
\end{figure*}


\bibliography{main}

\end{document}